\documentclass{article}
\usepackage{ctex}
\usepackage{PRIMEarxiv}

\usepackage[utf8]{inputenc} 
\usepackage[T1]{fontenc}    
\usepackage{hyperref}       
\usepackage{url}            
\usepackage{booktabs}       
\usepackage{amsfonts}       
\usepackage{nicefrac}       
\usepackage{microtype}      
\usepackage{lipsum}
\usepackage{fancyhdr}       
\usepackage{graphicx}       
\graphicspath{{media/}}     
\usepackage[export]{adjustbox}   
\usepackage{natbib}              
\usepackage[english]{babel}
\usepackage{polyglossia}
\usepackage{multirow}
\usepackage{subcaption}
\usepackage{array} 
\usepackage{lipsum} 
\usepackage{caption}
\usepackage{lastpage}
\usepackage{authblk} 
\usepackage{hyperref} 
\title{Mr3D-VL: A generalist vision–language foundation model for Multiparametric 3D Magnetic Resonance Imaging
}

\author{
Zhi Qiao$^{\dagger}$, Xintong Wu$^1$, Yichu He$^1$, Feng Shi$^1$
\\
$^1$ United Imaging Intelligence, Shanghai, China
\\
$\dagger$ Work completed while affiliated with Affiliation 1.
\\
mingshan\_ai@163.com
}

\date{} 

\begin{document}
\maketitle

\begin{abstract}
Multi-parametric magnetic resonance imaging (mpMRI) is a cornerstone for brain tumor diagnosis and treatment, yet current AI models face critical limitations: their lack of natural language interaction and interpretability impedes spatial information integration and cross-modal reasoning required clinically. Key challenges arise from significant physical meaning differences across modalities, spatial misalignment due to scan intervals, and the need for complex multi-feature interpretation in tasks like glioma grading. While visual-language models (VLMs) show promise in cross-modal understanding, existing methods focus mainly on 2D image modeling, neglecting direct perception of 3D volumetric space. Although 3D VLMs have been proposed for report generation and feature alignment in 3D CT imaging, mpMRI applications demand collaborative inference across multiple imaging modalities—a requirement unmet by current solutions. To address this, we introduce Mr3D-VL, a dedicated visual-language foundation model for multi-parametric 3D MRI. With 4 billion parameters, it employs an unsupervised pre-trained shared 3D encoder and 4D rotational positional embedding for dual modality-spatial integration. Its cross-modal projection layer uses a multi-resolution feature implantation strategy to enhance feature perception across resolutions. Experimental results show significant improvements over existing 4B/7B/30B domain-specific and general-purpose models in text generation tasks, achieving a BERTScore of 0.856 for report generation, with question-answering accuracy at 0.713 and multiple-choice accuracy at 0.912.

\end{abstract}



\section{Introduction}
Multiparametric magnetic resonance imaging (mpMRI) \cite{mpmri}, leveraging its advantages of multimodality, high soft-tissue contrast resolution, and non-invasiveness, plays an irreplaceable role in the diagnosis, grading, treatment planning, and therapeutic efficacy evaluation of brain tumors and other intracranial lesions. With advancements in medical imaging technology, clinical interpretation demands for mpMRI have evolved from simple tumor detection to clinically interpretable spatial information representation—requiring imaging analysis tools to not only accurately identify lesions but also support natural language-based interactive querying to assist neurosurgeons in critical tasks such as surgical planning, risk assessment, and patient-physician communication. However, current intelligent analysis models based on mpMRI exhibit significant limitations: clinicians cannot directly query tumor spatial characteristics, inter-modal associations, and clinical implications through natural language, thereby restricting their clinical applicability.

The primary challenges include: (1) substantial differences in tissue properties reflected by different modalities (e.g., T1, T2, DWI), necessitating models to understand the physical meanings and clinical values of each modality; (2) spatial misalignment caused by tissue displacement due to voluntary/involuntary patient motion across modalities with varying scan times; and (3) the need for cross-modal associative reasoning in clinical queries (e.g., "Does the T2 hyperintense region correspond to restricted diffusion on DWI?"), requiring models to integrate multimodal features. For instance, WHO grading of gliomas requires combining T2-FLAIR mismatch signs, ADC values, and rCBV perfusion parameters; differentiation of brain metastases necessitates observing "ring enhancement" on post-contrast T1 imaging and DWI hyperintensity; while meningioma management demands evaluating T2 signal homogeneity, adjacent meningeal "dural tail" signs, and bone invasion. These diagnostic logic frameworks require integrating spatial information, signal intensities, morphological features, and functional parameters from multimodal imaging into interpretable clinical descriptions.

Several attempts have been made to adapt 2D visual-language models (VLMs) to 3D medical imaging scenarios. For example, medGemma \cite{medGemma} treats 3D data as sequential 2D slices, lacking explicit 3D spatial relationship modeling. Hulumed \cite{hulu_med} proposed a 3D VLM for CT imaging, achieving progress in that domain but failing to address multimodal collaborative diagnosis challenges in mpMRI. mpLLM \cite{mpLLM} employed feature pooling for multimodal feature collaboration, yet could not resolve spatial misalignment across modalities while suffering from substantial information loss and reduced data-specific feature representation due to pooling operations—all contributing to insufficient clinical interpretability.

To address these gaps, this study introduces Mr3D-VL, a foundational visual-language model designed specifically for multimodal 3D MRI scenarios. Our key contributions include:

\begin{itemize}
\item[1] LLMs-Driven Data Generation: Leveraging large language models (LLMs) to parse clinical reports and integrate inference results from well-established smaller models (e.g., segmentation, detection), this approach generates diverse multimodal image-text datasets for cross-modal learning, addressing challenges of data scarcity and semantic incoherence.

\item[2] Unsupervised Pre-training with Shared Encoder: A modality-agnostic 3D visual encoder is shared across all imaging modalities, with unsupervised pre-training implemented via the Dino-v2 architecture. This enables adaptive capture of both intra-modal features and inter-modal disparities while reducing parameter counts and enhancing cross-modal fusion capabilities.

\item[3] 4D Rotational Positional Encoding: Incorporating structural characteristics of 3D volumetric data, a "depth" dimension is introduced to traditional 3D positional encoding (time, width, height), constructing a 4D coordinate system (time, depth, width, height). Rotational positional embeddings facilitate dual-dimensional fusion of modality information and spatial localization, improving reasoning about complex spatial relationships.

\item[4] Multi-Resolution Image Feature Implantation Strategy: Low-, medium-, and high-resolution feature maps are synchronously extracted from the pyramid-style feature extraction architecture of the visual encoder. Within the cross-modal projection module, a Vision Transformer (ViT) architecture fuses features across resolutions, which are then progressively implanted into the language model's decoder layers at each resolution level. This achieves collaborative attention to both global and detailed features, enhancing adaptability to clinical scenarios.

\end{itemize}

The introduction of Mr3D-VL marks a paradigm shift in intelligent analysis of multiparametric magnetic resonance imaging (mpMRI) toward clinically interpretable interaction. By enabling natural language querying capabilities, the model can directly respond to clinicians' diagnostic questions, providing interpretable descriptions incorporating anatomical localization, signal characteristics, and functional parameters. This significantly enhances the efficiency and accuracy of clinical decision-making in diagnosis and treatment planning. Future research could further expand the model's querying capabilities (e.g., supporting multi-turn dialogue and proactive clarification) while integrating additional clinical data (e.g., genomic information, treatment history) to construct a more comprehensive intelligent assistance system for brain tumor management. Moreover, through multicenter clinical validation and the development of standardized datasets, widespread clinical adoption of Mr3D-VL can be facilitated, ultimately achieving a closed-loop transition from "radiological diagnosis" to "precision medicine".

\begin{figure}
  \centering
  \includegraphics[width=\textwidth]{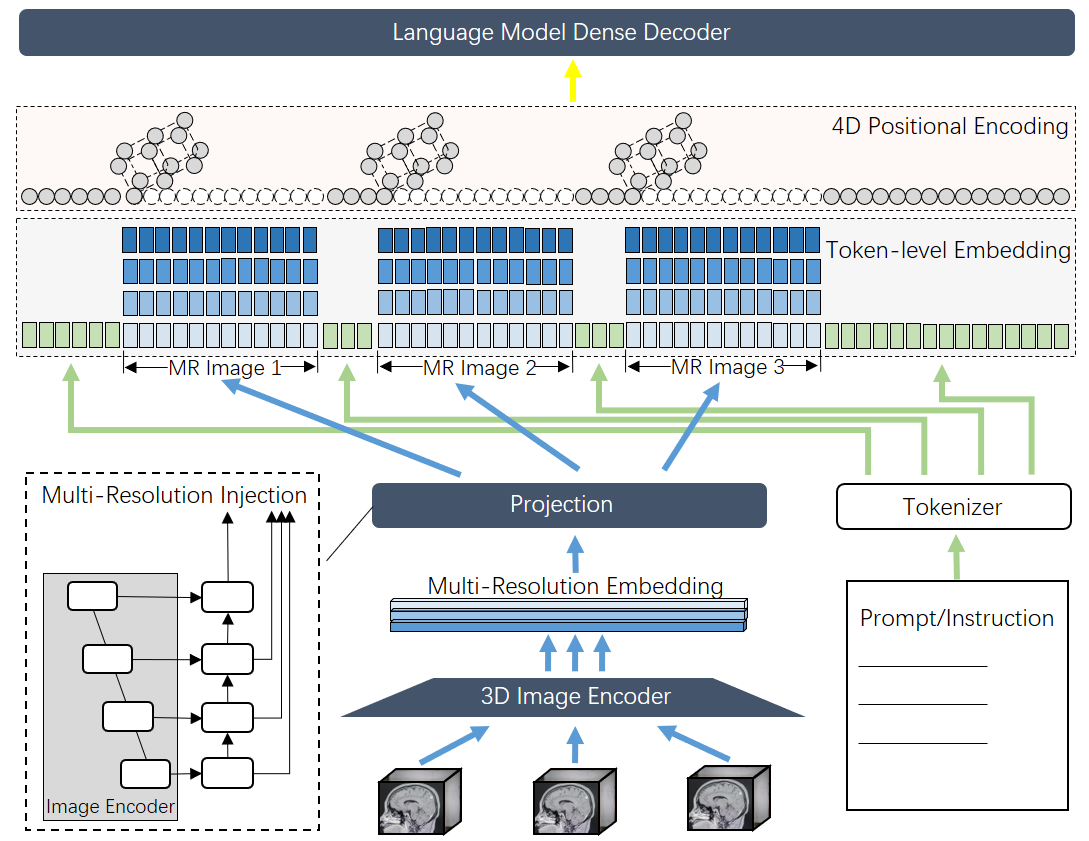}
  \caption{The Mr3D-VL framework integrates a vision encoder, a corss modality projection and a language model decoder to process multimodal MRI inputs, including T1, Flair, T2, ...}
  \label{fig:framework}
\end{figure}

\section{Related Works}
\subsection{Generalized‌ Vision Languagse}
The architectural evolution of vision-language models (VLMs) has transitioned from simple "concatenation" approaches to deeply "native unification" strategies. Leading model families exhibit significant divergence in the selection and fusion strategies of visual encoders, connectors (adapters), and language models, adapting to diverse requirements ranging from edge deployment to complex reasoning, and from static images to dynamic videos.

\textbf{Visual Encoder}~~The Vision Transformer (ViT) has been widely adopted as the visual encoder in mainstream VLMs, such as the InternVL series \cite{internvlvl}, Qwen series \cite{qwen3vl}, Seed 1.5 VL \cite{seed15vl}, and Kimi-VL \cite{kimik25}. For ViT pre-training, contrastive learning strategies \cite{clip} are typically employed for initial visual encoder training. In contrast to these dominant VLM architectures, our approach utilizes a CNN-based framework \cite{cnn_3d} as the visual encoder, leveraging the DINO framework \cite{dinov2} for unsupervised learning. In medical imaging (e.g., CT, MRI, pathology slides), while ViT demonstrates strong capabilities in natural image processing, CNNs remain the dominant backbone in both clinical applications and academic research due to several advantages:(1) Translation Invariance and Localization Assumptions: CNNs inherently incorporate translation invariance and locality assumptions, enabling faster convergence even on small datasets while mitigating overfitting risks \cite{cnn_consistent,cnn_smalldata}. (2) Computational Efficiency: Convolutional operations exhibit O(N) complexity and efficiently handle ultra-large images through sliding windows and multi-scale strategies, ensuring controlled GPU memory usage \cite{cnn_time}. (3) Hierarchical Feature Extraction: Convolutional kernels focus on local feature extraction, with lower layers capturing edges/textures and higher layers extracting semantic information. This hierarchical structure aligns with human visual perception and medical image interpretation logic \cite{cnn_local}. (4) 3D Spatial Continuity: 3D CNNs (e.g., 3D U-Net) perform volumetric convolutions directly in voxel space, fully exploiting z-axis continuity information \cite{cnn_3d}. Contrastive learning frameworks like CLIP \cite{clip} rely on massive "image-text" pairs to align visual and textual semantics. However, high-quality "image-text" datasets are extremely scarce in medicine, particularly for multiparametric MRI scenarios where clinicians provide holistic judgments across multimodal imaging rather than textual descriptions for single modalities. The prohibitive costs and expertise requirements of data annotation have long hindered large-scale supervised learning applications. DINO's purely visual self-supervised learning (SSL) \cite{dinov2} eliminates dependence on textual labels entirely. Through a "student-teacher" network architecture, it enables models to learn intrinsic image structures, textures, and shape invariances autonomously, addressing medical imaging's "data silos" and "annotation bottlenecks."

\textbf{Adapter Layer}~~LLaVA \cite{llava} simply concatenates visual features at the language model's input layer, while PaliGemma \cite{paligemma} employs a linear layer (instead of complex MLPs or Transformers) as the connector. Dynamic routing attention mechanisms or cross-attention layers have also been explored for adapter design \cite{kimivl,glm}. Qwen3-VL \cite{qwen3vl} introduces DeepStack technology \cite{deepstack}, dynamically injecting visual features into multiple LLM layers (rather than solely at the input layer) to achieve deep vision-language fusion. DeepSeek VL \cite{deepseek-vl} adopts a hybrid visual encoder combining a "semantic-aligned encoder" (for global semantics) and a "high-resolution encoder" (for fine details), fused via specialized projection layers to support high-resolution learning. MiniCPM-V \cite{MiniCPM_V} replaces traditional MLPs with a Resampler module, compressing visual features into a fixed number of tokens while enabling unified tokenization (single/multi-image/video processing) and standardized workflows. While these models primarily train on natural images (with some incorporating medical imaging), natural images typically feature clear object boundaries and high foreground-background contrast, facilitating semantic disambiguation. In medical imaging, many lesions (e.g., early-stage liver cancer, white matter lesions) exhibit minimal grayscale differences from surrounding tissues, requiring diagnosis based on subtle textural variations (e.g., lung nodule spiculation, nuclear atypia in cell membranes) at sub-pixel levels. Although resampling techniques significantly reduce feature dimensionality, they risk losing critical detail information. Drawing inspiration from DeepSeek VL's high-resolution encoder performance and Qwen3-VL's successful DeepStack application, we propose a specialized multi-resolution visual token implantation strategy for medical imaging. This approach fully leverages features generated at different resolutions by the visual encoder to achieve deep vision-language fusion while preserving diagnostic-relevant details.

\begin{figure}
  \centering
  \includegraphics[width=\textwidth]{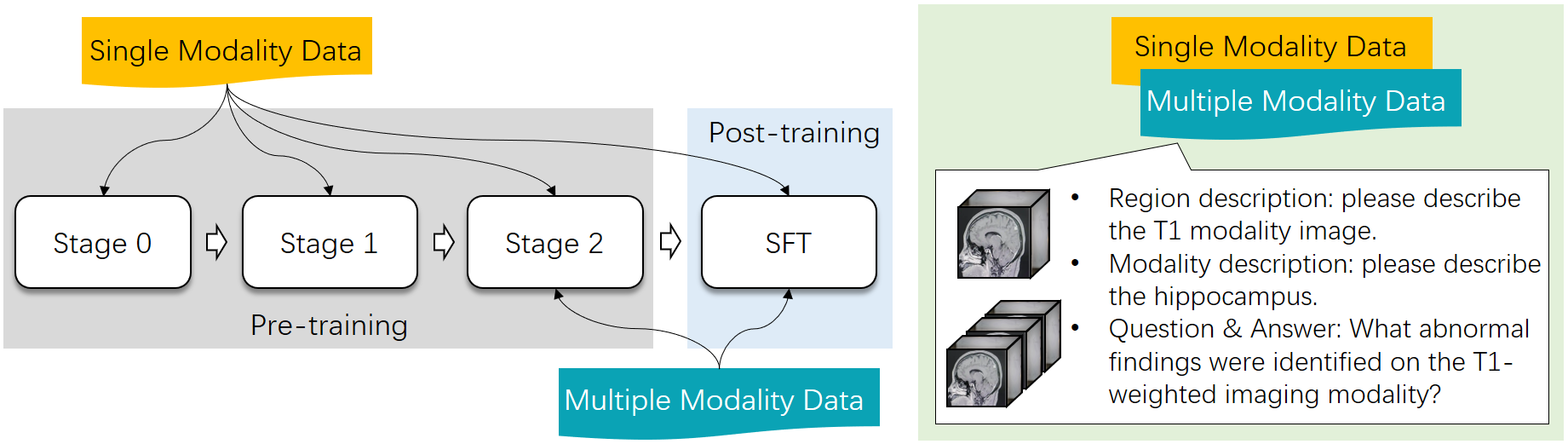}
  \caption{The pipeline of Pre-training \& Post-training}
  \label{fig:work_flow}
\end{figure}
\subsection{Vision Language in Medical Domain}
LLaVA-Med \cite{llava-med} rapidly transforms general-purpose vision-language capabilities into medical expertise through high-quality medical instruction tuning \cite{instructiontuning}. With relatively limited modality support primarily focused on 2D imaging, its training data is sourced from authoritative repositories like PubMed\footnote{https://pubmed.ncbi.nlm.nih.gov/}, utilizing 2D illustrations from biomedical literature. Its streamlined architecture employs a simple adapter to bridge the visual encoder and large language model (LLM), lowering barriers for secondary development.MedGemma \cite{medGemma} achieves accurate identification of X-rays, CT scans, and pathology slides—even supporting 3D volumetric processing—by integrating a visual encoder trained via the MedSigLIP strategy and adopting Gemma3 \cite{gemma3} as its LLM backbone. Its architecture resembles LLaVA-Med’s simple adapter-based image-text concatenation approach, a strategy also employed by Lingshu \cite{lingshu}. Both MedGemma and Lingshu propose detailed dataset processing pipelines, constructing massive multimodal datasets encompassing text and imaging while innovatively incorporating Long Chain-of-Thought (Long CoT) reasoning \cite{longcot} to enhance inference quality. Unlike models that flatten 3D slices or sample video frames as compromises, Hulu-Med \cite{hulu_med} treats image patches as universal units through 2D Rotary Position Embedding (2D RoPE). This enables it to analyze hour-long surgical videos or 3D CT volumes by comprehending both anatomical structures and the temporal logic of instrument movements. However, despite its 2D RoPE innovation, Hulu-Med’s position encoding remains independent of text tokens. The concatenated token sequence still relies on 1D RoPE \cite{rope}, preventing contextual interactions from perceiving spatial relationships among visual tokens. Additionally, Hulu-Med processes 3D data as stacked 2D slices, where (1) Disrupts volumetric spatial features by ignoring inter-slice correlations; (2) Increases token count exponentially, straining GPU memory; (3) Requires larger-scale training data due to prolonged context windows. Unlike existing medical VLMs, our method directly models 3D volumetric data while supporting multi-volume training and inference. We propose 4D Rotary Position Embedding (4D-RoPE), which assigns distinct position encodings to: Text tokens (1D sequence), Volumetric depth (z-axis), Height (y-axis) and Width (x-axis). This unifies positional information across modalities within the LLM’s decoding space, enabling bidirectional capture of spatial relationships during vision-language interactions. By preserving intrinsic 3D structural integrity while reducing computational overhead, our approach addresses key limitations of prior works in medical imaging analysis.

\subsection{Dataset for 3D Vision Language Model}
In the development of vision-language foundation models, the curation of pretraining and fine-tuning datasets serves as the cornerstone determining both the upper bound (performance ceiling) and lower bound (application reliability) of model capabilities, particularly in specialized fields like healthcare. While numerous open-source medical datasets exist for training and evaluating vision-language models, these predominantly focus on 2D imaging modalities \cite{data2d-1,data2d-2,data2d-3,data2d-4}. Although some efforts have constructed large-scale medical multimodal datasets incorporating computed tomography (CT), positron emission tomography (PET), magnetic resonance imaging (MRI), and other modalities, the majority of these data are derived from book illustrations or internet screenshots. Despite involving 3D modalities in theory, the selected images often represent single 2D slices from 3D volumetric data \cite{data2d-5,data2d-6,data2d-7}. The acquisition of 2D image-text datasets is relatively straightforward—scalable through web scraping or book digitization—resulting in massive datasets. In contrast, 3D image-text datasets, which better align with clinical requirements, face significant collection challenges due to privacy regulations, annotation complexity, and storage demands. Consequently, existing 3D medical datasets are limited in scale and predominantly feature CT modality \cite{data3d-1,data3d-2,data3d-3,data3d-4}. Current approaches to 3D representation learning and vision-language model training rely on unimodal cross-modal alignment (e.g., Hulumed series \cite{hulu_med} supports 3D reasoning but uses CT-exclusive training data). However, real-world clinical scenarios demand multi-modal 3D imaging inference (e.g., joint interpretation of multi-parametric MRI sequences), a challenge that remains unaddressed by existing solutions. Furthermore, no publicly available datasets support both pretraining and fine-tuning for such multi-modal 3D tasks.

To bridge this gap, our work introduces a large language model (LLM)-powered synthetic data generation pipeline comprising four key components: (1) Radiology report parsing to extract structured clinical information; (2) Expert-guided segmentation/detection annotation integration for precise spatial grounding;
Textual data augmentation via medical knowledge-aware paraphrasing; (3) Question-answer pair simulation to generate diverse diagnostic reasoning samples. This framework systematically addresses the scarcity of multi-modal 3D medical vision-language datasets, enabling end-to-end training of models capable of complex volumetric reasoning while maintaining clinical interpretability.

\subsection{Position Encoder}
The permutation invariance of self-attention mechanisms necessitates positional encoding to inform large language models (LLMs) about sequence order, relative distances, and structural dependencies. While early methods relied on absolute positional embeddings \cite{attention}, relative positional encoding, which generalizes better to varying sequence lengths, has become the standard. Notably, Rotary Positional Embedding (RoPE) \cite{roformer} has emerged as the de facto choice for modern LLMs (e.g., Llama \cite{llama3} and Qwen \cite{qwen}). Vision-language models (VLMs) require modality-aware positional encoding to handle heterogeneous inputs, including 1D text and 2D/3D visual data. Current approaches fall into two categories: 1D sequential designs (e.g., vanilla RoPE \cite{roformer} and V2PE \cite{v2pe}), which flatten and concatenate all inputs into a single sequence. While simple, this approach discards native visual geometry, leading to significant performance drops in tasks requiring visual localization and spatial reasoning. Multi-dimensional designs, which extend RoPE across multiple axes (e.g., time, height, width) by partitioning embedding channels. For instance, Qwen2-VL \cite{qwen2-vl} introduces Multimodal RoPE (MRoPE) to unify positional encoding for text and visual tokens. However, MRoPE allocates positional embeddings to t-h-w blocks, placing temporal information entirely in high-frequency channels.
While general-purpose VLMs primarily process 2D natural images (e.g., extending to multi-view or video scenarios), medical imaging predominantly involves 3D volumetric data (e.g., CT, PET, MRI). When adapting VLMs to medical contexts, some works repurpose MRoPE by treating 3D volumes as sequences of 2D slices \cite{hulu_med,medGemma}. However, this approach disrupts the spatial integrity of volumetric data and inevitably causes an exponential surge in token count, complicating training due to exploding memory requirements and data scalability issues. To address these limitations, we propose a 4D Rotary Positional Encoding (4D-RoPE) strategy tailored for 3D volumetric data, extending positional axes to (time, depth, width, height). This approach preserves native 3D spatial relationships while enabling efficient cross-modal alignment in medical VLMs.

\section{Model Architecture}
Following the typical vision language framework, Mr3D-VL adopts a three-module architecture comprising a vision encoder, an MLP-based vision–language projection, and a large language model (LLM). Figure \ref{fig:framework}
depicts the detailed model structure.

\textbf{Large Language Model} In the primary configuration of Mr3D-VL, the Qwen3-4B-Instruct\cite{qwen3} serves as the foundational language model backbone. Textual embeddings and projected 3D visual embeddings are concatenated to form a cohesive input sequence. The model then processes this sequence autoregressively, predicting subsequent tokens based on preceding visual and textual cues. This architecture endows Mr3D-VL with the versatility to undertake a variety of generative tasks without requiring task-specific modifications. For textual input processing, we utilize the Qwen2Tokenizer\cite{qwen2}, an integral component of the large language model's backbone. This tokenizer, grounded in the Byte-Pair Encoding (BPE\cite{bpe}) algorithm, boasts a substantial vocabulary size of 151,936 tokens, facilitating nuanced and efficient text representation within the model.

\textbf{3D Vision Encoder} 
The visual encoder stands as a pivotal component within visual-language models, necessitating pre-training before integration into the overarching cross-modal visual-language framework. A prevalent pre-training methodology is contrastive learning, which hinges on a substantial corpus of image-text description pairs. Magnetic resonance (MR) imaging encompasses dozens of modalities, with over 10 commonly utilized in clinical settings, such as T1-weighted imaging, T2-weighted imaging, FLAIR, DWI, DTI, SWI, MRS, fMRI, and PWI. Physicians select various modalities based on the symptoms and diagnostic hypotheses, synthesize the modal images, and generate imaging reports. Consequently, each case often involves multiple modalities of data, presenting a significant challenge of extensive missing annotated data in such scenarios. Previous studies\cite{merge_mp,concat_mp} have adopted multimodal merging strategies, fusing and compressing multimodal images and utilizing comprehensive imaging reports as textual supervision signals for encoder pre-training via contrastive learning. However, this compression approach tends to emphasize holistic features while overlooking individual modality-specific characteristics. Moreover, employing comprehensive reports as alignment labels introduces semantic complexity and confusion, hindering the learning of visual features. mpLLM \cite{mpLLM} introduces independent feature extraction pathways (mixture of experts) for each modality, substantially increasing parameter size and leading to encoder decoupling. This results in information fragmentation across modalities and neglects potential inter-modal correlations. Drawing inspiration from the universal encoder approach in 2D visual-language models, we advocate for a modality-agnostic 3D imaging encoder shared across all modalities. This enables the model to capture both commonalities and differences among diverse modalities. We select MedNext \cite{MedNeXt} as the backbone network for the visual encoder, leveraging its innovative designs such as the ConvNeXt3D architecture \cite{convnextv2}, UpKern technology, and composite scaling strategies. These advancements facilitate long-range spatial dependency capture, cross-scale semantic information preservation, and performance breakthroughs in data-limited scenarios when processing three-dimensional volumetric data. Employing the Dino-v2 architecture for unsupervised pre-training, our approach eliminates the need for textual supervision, focusing instead on perceptual feature capture. Higher-dimensional semantic feature extraction is delegated to the projection layer and large language model. This design not only reduces model parameters but also enhances cross-modal feature fusion capabilities, enabling the model to comprehend the intrinsic relationships among multimodal images during the pre-training phase.

\textbf{Multimodal 4D Rotary Position Embedding}.~~~Positional embeddings are crucial for modeling sequential data in both vision and language modalities. Building upon the Multimodal Rotary Position Embedding (MRoPE) introduced in Qwen2-VL, we extend its capabilities to better handle spatial information in 3D volume data and proposed the Multimodal 3D Rotary Position Embedding (M3RoPE). The M3RoPE in Mr3D-VL decomposes the position embedding into four distinct components: temporal, depth, height, and width to effectively model multimodal inputs. For textual inputs, all four components use identical position IDs, making M4RoPE functionally equivalent to traditional 1D RoPE \cite{roformer}. For 3D volume data, the temporal ID remains constant across visual tokens, while unique IDs are assigned to the height and width components based on each token’s spatial position within the volume data. 

\textbf{Multi-resolution Vision Token Injection}. Drawing inspiration from DeepStack \cite{deepstack}, we propose a novel approach that integrates visual tokens into the multi-layer architecture of a large language model (LLM). While the original DeepStack method stacks tokens derived from multi-scale visual inputs, and Qwen3-VL extends this by extracting visual tokens from intermediate layers of the visual Vision Transformer (ViT), our approach diverges significantly to address the unique demands of medical imaging. In medical imaging, varying resolutions hold distinct significance for identifying different types of lesions and reaching a final diagnosis. Unlike the aforementioned methods, we leverage output features of different resolutions from the visual encoder. These features are then incorporated into the intermediate layers of the ViT within the projection module through skip connections—a design inspired by the UNet decoder \cite{unet}. This strategy aims to progressively enrich the feature learning process with spatial information. Subsequently, tokens are extracted from the intermediate layers of the ViT and injected into the multi-layer structure of the language model. 

\begin{figure}
    \centering
    \begin{subfigure}[b]{0.45\textwidth}
        \includegraphics[width=\textwidth]{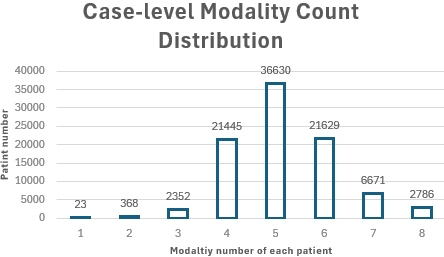}
        \caption{for Different input modality number in the dataset}
        \label{fig:modality_count_distribution}
    \end{subfigure}
    \hfill
        \begin{subfigure}[b]{0.45\textwidth}
        \includegraphics[width=\textwidth]{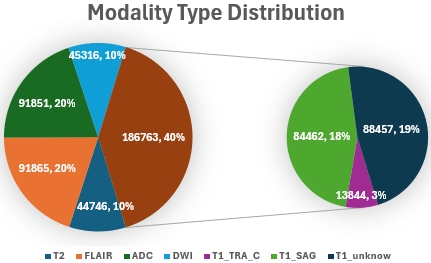}
        \caption{for Different modality types in the dataset}
        \label{fig:modality_type_distribution}
    \end{subfigure}
    \caption{Data distribution of original mpMRI data}    \label{fig:data_distribution}
\end{figure}
\section{Training Pipeline}
The entire training process is divided into two main stages: pre-training and post-training. The pre-training phase is further subdivided into three sequential sub-steps, progressing from simple to complex data and from partial to full model parameter updates. During the post-training phase, we conduct goal-oriented supervised fine-tuning (SFT), as illustrated in Figure \ref{fig:work_flow}. We have collected a dataset comprising 103,605 cases of brain multiparametric magnetic resonance imaging (mpMRI), totaling 460,541 images. On average, each case includes 4.4 different modality images. Statistical information regarding the distribution of different modality types per case is shown in Figure \ref{fig:modality_count_distribution}, while the overall statistics for each of the eight modalities involved are presented in Figure \ref{fig:modality_type_distribution}. 

All image data undergoes preprocessing as follows: First, linear interpolation resampling is applied to achieve a voxel space of $1×1×5~mm^3$, a choice motivated by the prevalence of thick-slice MRI in clinical multiparametric settings. Subsequently, the data is cropped to dimensions of [32, 192, 192]. Finally, pixel values are normalized to the range of -1 to 1. Regarding the imaging reports, detailed processing methods will be elaborated upon in both the pre-training and post-training sections. To facilitate a comprehensive evaluation of model performance, we partition the dataset by case (patient), ensuring that all test data remains entirely unseen during both the pre-training and post-training phases. In the following sections, we will provide detailed descriptions of the pre-training and post-training procedures.

\subsection{Pre-Training}
\subsubsection{Training Recipe}
To enable the model to evolve from having single-modal short-text alignment capabilities to possessing mixed-modal long-context understanding abilities, we systematically divide the pre-training strategy into three distinct phases.

\textbf{Stage S0: Visual-Language Alignment} This phase focuses on efficiently bridging the modality gap between the visual encoder and the large language model (LLM). During this stage, only the parameters of the projection module are trained, while both the visual encoder and the LLM backbone remain frozen. We utilize a unimodal dataset comprising approximately 1 billion tokens, consisting of image-text pairs characterized by single-modal short-text attributes, including modality descriptions, local descriptions, and question-and-answer (Q\&A) pairs.

\textbf{Stage S1: Unimodal Pre-training}
In this phase, we unfreeze all model components—the visual encoder, projection module, and LLM—for joint end-to-end training. The same unimodal dataset of approximately 1 billion tokens used in Phase S0 is repurposed for this stage.

\textbf{Stage S2: Multimodal Pre-training} This phase aims to significantly expand the model's contextual processing capabilities. A key change in this stage is the transition from single-modal to mixed-modal inputs, with all model parameters participating in the training process. Training is conducted on a mixed-modal dataset containing approximately 1 billion tokens. This phase is crucial for enabling the model to handle and reason over mixed-modal inputs effectively.

\subsubsection{Pre-Training Data}
Below, we will delve into the data collection efforts for medical multi-parametric magnetic resonance imaging (MRI) from four distinct dimensions: modality description-based, tissue/organ-based, grounding-based, and spatial understanding-based approaches. For each category of data, we further divide the construction process into two subcategories: unimodal data construction and mixed-modal data construction. Detailed specifics regarding the pre-training stage data are presented in Table \ref{tab:pretrain_data_distribution}.

\textbf{Data Construction based on Modality Description}~~
To construct unimodal image-text pairs by obtaining textual descriptions corresponding to each modality, we leverage the powerful text comprehension capabilities of large language models (LLMs). Initially, we dissect the original imaging reports, separating content related to each modality based on its distinct characteristics. Subsequently, the segmented textual content undergoes further refinement by the LLM, which rewrites it to enhance logical coherence and semantic richness. Since LLMs are typically pre-trained on medical corpora, the rewriting process naturally incorporates explanations of medical terminology and elucidates the logical relationships between modality-specific symptoms and final diagnoses. This text reconstruction not only transforms concise and often obscure imaging reports into logically coherent and semantically enriched narratives but also introduces external knowledge. For instance, in the case of T1-weighted images, the LLM extracts textual sentences containing the term "T1" from the imaging report. It then integrates and restates highly relevant content, such as differences in signal intensity across various tissues, precise delineation of boundaries between cerebral gray and white matter, and characteristics of potential lesion regions. This approach ensures accurate and comprehensive interpretation of unimodal content. Such descriptions facilitate a more intuitive understanding of image information for clinicians while providing rich textual annotations for subsequent data analysis and model training.

Furthermore, from the structured imaging reports, we randomly select multiple modalities along with their corresponding descriptions. Using the LLM, we then rewrite these descriptions into a cohesive multimodal imaging narrative. This method allows the generation of multiple distinct experimental samples from a single case while eliminating instances where a modality is mentioned in the original report but absent from the imaging data, thereby preventing informational confusion. Additionally, it expands the report to include detailed descriptions across various modalities. For example, a report might mention findings from both T1- and T2-weighted images. Through LLM reasoning, information from both modalities is integrated to provide a detailed description of the lesion's manifestations across different modalities, including its size, shape, and signal characteristics. This enriched presentation of mixed-modal data offers clinicians a more comprehensive and accurate basis for diagnosis.


\begin{table}[]
\centering
\begin{tabular}{lrrr}
\hline
           & \multicolumn{1}{l}{Stage-0} & \multicolumn{1}{l}{Stage-1} & \multicolumn{1}{l}{Stage-2} \\ \hline
           Dataset     & 2711712                     & 936404                      & 406130                      \\ \hline
\end{tabular}
\caption{Data details of Pre-training stage}
    \label{tab:pretrain_data_distribution}
\end{table}

\begin{table}[]
\centering
\begin{tabular}{cccc}
\hline
                          &        & Training        & Testing       \\ \hline
\multirow{2}{*}{Q\&A Task} &  Open-ended question-answer & 100108     & 10000     \\ \cline{2-4} 
                          & Multiple-choice questions & 100000     & 10000     \\ \hline
Text Generation Task                    & Radiology report generation   & 88962      & 7959      \\ \hline
\end{tabular}
\caption{Data details of Post-training stage}
    \label{tab:post_data_distribution}
\end{table}

\textbf{Data Construction based on Anatomy}~~We begin by dissecting the original imaging reports, segregating the content related to different anatomical organs and regions based on their distinct characteristics. Subsequently, we employ a large language model (LLM) to rewrite the text, generating logically coherent and semantically rich descriptions. For instance, in brain imaging reports, if there is a description pertaining to the hippocampus, the LLM can further supplement the text with background knowledge regarding the morphology, volumetric changes, and relationships with surrounding tissues of the hippocampus. It then integrates this supplementary information with the existing content in the report to provide a re-described account of this region. Our objective is to foster stronger cross-modal learning between the textual descriptions of different anatomical regions and their corresponding image areas. We aim for the model to perceive distinct image regions when analyzing the textual descriptions of various anatomical parts. To achieve this, for each case sample, we require multi-regional image-text data, compelling the model to perform matching and discrimination tasks between the textual descriptions of different anatomical parts and their corresponding image regions. This approach enhances the model's ability to integrate and interpret multimodal medical data effectively.

\textbf{Data Construction based on Segmentation\&Detection results}~~
Small models exhibit unparalleled performance advantages in single-task scenarios, such as classification, detection, and segmentation. Therefore, we propose a data generation scheme based on the results derived from small models. In our dataset, a subset of data includes organ segmentation results alongside lesion detection outcomes. By leveraging these tissue structure segmentation results and lesion detection findings, we can further deduce the spatial relationships between lesions and anatomical organs. Utilizing a large language model (LLM), we describe the local regions from the perspectives of both tissue structures and lesions. For instance, in thoracic magnetic resonance imaging (MRI), given the known positions and morphologies of organs such as the heart and lungs, along with the localization of potential lesions like pulmonary nodules or cardiac valve abnormalities, the LLM can integrate the positional relationships between organs and lesions. It then describes the interplay between the lesions and normal organs, supplementing the concepts of organs and lesions, as well as their relationships, with the LLM's inherent medical knowledge. This approach aids clinicians in comprehending the biological structures within a single modality, thereby enhancing diagnostic accuracy.

For mixed-modality data, we randomly select multiple modalities, multiple anatomical organs, and multiple lesion regions. Subsequently, we employ the LLM to generate more complex descriptions. In the generated textual descriptions, nouns referring to anatomical organs and abnormal regions are highlighted by enclosing their corresponding bounding boxes within and tags, respectively. Additionally, the LLM supplements the descriptions with bounding boxes delineating the regions of tissue structures and lesions.

\textbf{Data Construction based on Spatial Cognition}~~By reusing the aforementioned segmented imaging report data, which has been partitioned according to modality and/or anatomical regions, and incorporating partial grounding results, we aggregate all the information and input it into a large language model (LLM) to generate question-and-answer (Q\&A) datasets. Furthermore, by adjusting the prompts, we compel the LLM to produce closed-set Q\&A questions with predefined options. For instance, given a brain magnetic resonance imaging (MRI) scan, the LLM can generate questions such as "In which cerebral lobe is the lesion located?" and "What is the signal intensity of the lesion?", along with their corresponding answers. This approach not only facilitates the evaluation and enhancement of the model's understanding of spatial information within a single modality but also provides clinicians with a rapid query tool and diagnostic aid.

By conducting data collection and organization efforts across the aforementioned four dimensions, we can establish a rich, accurate, and comprehensive data foundation for the development of visual-language models tailored to multi-parametric magnetic resonance imaging (MRI) in the medical domain.

\subsection{Post-Training}
The post-training pipeline is designed to refine the model's instruction-following capabilities. Our primary objective is to equip the model with the capability to precisely handle tasks related to 3D multimodal magnetic resonance (MR) imaging across diverse and complex medical scenarios. This includes the ability to articulate and infer overall imaging features and potential disease risks from 3D MR images, as well as to perform comprehensive reasoning by correlating the representational characteristics of multimodal images. 


\subsubsection{Post-Training Data}
 Guided by these target tasks, we systematically constructed a Supervised Fine-Tuning (SFT) dataset comprising nearly 300,000 samples. This dataset is composed of 3D MR imaging report generation data and Question-and-Answer (Q\&A) data. The Q\&A data is further categorized into open-ended questions and multiple-choice questions, with the latter directly reusing the Q\&A dataset from the pre-training phase. Detailed data distributions are presented in Table \ref{tab:post_data_distribution}.

For the report generation dataset, we leverage a Large Language Model (LLM) to expand and supplement original medical imaging reports, significantly enhancing their completeness and semantic coherence. For instance, in brain imaging reports, if there is a description involving the hippocampus, the LLM can further enrich the report with background knowledge on the hippocampus's morphology, volume changes, and its relationship with surrounding tissues. Additionally, through prompt engineering, the LLM automatically identifies anatomical region information mentioned in the reports, such as "left upper lobe of the lung" or "right lobe of the liver," and structurally annotates them with tags. Simultaneously, for descriptions of abnormal lesions, such as key pathological features like "ground-glass nodules" or "calcifications," embedded tags are used to achieve visual highlighting. This dual-tagging system not only makes regional localization and abnormal identification more intuitive but also strengthens the logical chain of the report through semantic associations.

\section{Evalution}
We conducted a systematic assessment of Mr3D-VL's performance in interpreting 3D medical images across two tasks: Visual Question Answering (VQA) and Medical Report Generation (MRG). The VQA task was further subdivided into two categories: open-ended questions and multiple-choice questions.

\subsection{Settings}
The test data primarily originated from the fine-tuning dataset constructed during the post-training phase, with detailed statistics provided in Table \ref{tab:post_data_distribution}. All experiments were conducted using 4 NVIDIA A40 GPUs. For comprehensive benchmarking, we evaluated our model against medical vision-specific models (e.g., Lingshu \cite{lingshu}, HuluMed \cite{hulu_med}) and general-purpose models (e.g., Qwen3.5 series \cite{qwen3vl}). Since Lingshu and the Qwen3.5 series natively lack support for 3D volumetric data, we adapted them for 3D evaluation by slicing each volume into sequential 2D image stacks and treating the task as a multi-image assessment. While HuluMed supports volumetric data, its training corpus predominantly consisted of CT scans, which differ significantly from MRI in imaging principles and visual characteristics. Thus, we uniformly adopted a multi-image evaluation protocol for 3D tasks.

We employed the following five metrics to assess model performance:
\begin{itemize}
\item[*] BLEU-4 \cite{bleu}: Evaluates fluency and local accuracy of generated text.
\item[*] ROUGE-L \cite{rouge}: Measures coherence and information completeness in generated text.
\item[*] BERTScore \cite{bertscore}: Assesses semantic plausibility using the multilingual MiniLM embedding model\footnote{https://hf-mirror.com/sentence-transformers/paraphrase-multilingual-MiniLM-L12-v2}.
\item[*] METEOR \cite{meteor}: Gauges lexical diversity and syntactic flexibility of generated text.
\item[*] CIDEr \cite{cider}: Evaluates naturalness and human preference in generated text.
\item[*] RadGraph-F1 \cite{radgraphf1}: Measures their ability to accurately capture and represent clinical entities and relationships.
\end{itemize}

\subsection{Efficiency and Deployability Analysis}
We conduct the FLOPs and max GPU memory utilization computation for the proposed MR3D model, along with comparisons to the mainstream large models. The details are shown in Table.~\ref{tab:flops}.

For Computational Efficiency (FLOPs), The MR3D model demonstrates significantly lower FLOPs compared to mainstream large models (e.g., Qwen3.5-4B, Hulu-med-7B, Lingshu-7B) under both unimodal and 5-modal input settings. For 5-modal input, MR3D achieves ~96\% reduction in FLOPs (2,064.64 G vs. 57,186.88 G for Qwen3.5-4B) while maintaining competitive performance, indicating superior efficiency in multimodal processing. For GPU Memory Utilization, MR3D exhibits minimal GPU memory consumption across all configurations, with a peak usage of 9,215.77 MB under 5-modal input-~93\% lower than Lingshu-7B (115,741.43 MB) and ~86\% lower than Hulu-med-7B (66,181.22 MB). This lightweight footprint makes MR3D highly scalable for resource-constrained environments (e.g., edge devices, low-end GPUs).

Despite its smaller parameter size (4B vs. 7B for Hulu-med/Lingshu), MR3D achieves comparable or better efficiency in multimodal tasks, suggesting optimized architecture design (e.g., sparse attention, cross-modal fusion mechanisms). The model’s low FLOPs and memory usage align with real-world deployment needs, particularly in clinical settings where rapid inference and cost-effectiveness are critical.

\begin{table}[]
\scriptsize
\centering
\begin{tabular}{|l|c|r|rr|rr|}
\hline
\multirow{2}{*}{Model} & \multicolumn{1}{l|}{\multirow{2}{*}{Model Version}} & \multicolumn{1}{l|}{\multirow{2}{*}{Parameter Size (M)}} & \multicolumn{2}{c|}{1-modal Input}                       & \multicolumn{2}{c|}{5-modal Input}                       \\ \cline{4-7} 
                       & \multicolumn{1}{l|}{}                               & \multicolumn{1}{l|}{}                                    & \multicolumn{1}{r|}{Max GPU Utilization (M)} & FLOPs (G) & \multicolumn{1}{r|}{Max GPU Utilization (M)} & FLOPs (G) \\ \hline
Qwen3.5                & 4B                                                  & 4536.73                                                  & \multicolumn{1}{r|}{9948.56}                 & 11481.12  & \multicolumn{1}{r|}{14447.93}                & 57186.88  \\ \hline
Hulu-med               & 7B                                                  & 7499.54                                                  & \multicolumn{1}{r|}{21076.72}                & 47735.32  & \multicolumn{1}{r|}{66181.22}                & 121409.3  \\ \hline
Lingshu                & 7B                                                  & 7201.88                                                  & \multicolumn{1}{r|}{38296.57}                & 15423.61  & \multicolumn{1}{r|}{115741.43}               & 73770.58  \\ \hline
Mr3D                   & 4B                                                  & 4048.12                                                  & \multicolumn{1}{r|}{8726.71}                 & 572.07    & \multicolumn{1}{r|}{9215.77}                 & 2064.64   \\ \hline
\end{tabular}
\caption{FLOPs and GPU Memory Utilization}
\label{tab:flops}
\end{table}

\subsection{Experimental Results}
When assessed on medical imaging report generation tasks, the QWen3.5 series models outperformed two dedicated medical imaging foundation models—Hulu-Med and Lingshu—across multiple metrics, including the smaller-scale QWen3.5-4B variant. Report generation represents a multimodal generation challenge, requiring models to interpret 3D imaging data from multiple modalities while managing the cognitive load imposed by vast volumes of visual information. Although the QWen series also faces similar challenges, its native multimodal training approach enhances image comprehension capabilities. Among QWen3.5 variants, QWen3.5-27B demonstrated superior performance over QWen3.5-35B on BLEU-4, ROUGE-L, BERTScore, and METEOR, suggesting that dense architectures achieve higher accuracy than mixture-of-experts (MoE) models at comparable parameter scales. Notably, Mr3D-VL surpassed all competing models, achieving marked improvements across all metrics: BLEU-4 (0.169) and ROUGE-L (0.353). Given that our experiments utilized a non-public dataset with unique preprocessing protocols, metrics like BERTScore, METEOR, and CIDEr better reflect Mr3D-VL’s advantages in handling multi-parametric MRI (mpMRI) data. Compared to existing models, Mr3D-VL achieved over 20\% improvements across these three evaluation benchmarks: BERTScore (0.856), METEOR (0.496), and CIDEr (0.655). Detailed results are provided in Table \ref{tab:report_generation_accuracy}.

\begin{table}[]
\centering
\begin{tabular}{llllllll}
\hline
\multicolumn{8}{c}{Report Generation}                                                                                       \\ \hline

                 & Size   & BLEU-4         & ROUGE-L        & BERTScore & METEOR         & CIDEr     & RadGraph-F1 \\ \hline
                 & \multicolumn{6}{c}{General-purpose Multimodal VLMs}                                         \\ \hline
QWen3.5          & 4B     & 0.022          & 0.154          & 0.688          & 0.379          & 0.444    &   0.555   \\
QWen3.5          & 27B    & 0.031          & 0.178          & 0.733          & 0.400          & 0.452    &   0.580   \\
QWen3.5          & 35B    & 0.024          & 0.163          & 0.702          & 0.374          & 0.481    &   0.399   \\ \hline
                 & \multicolumn{6}{c}{Medical Multimodal VLMs}                                                 \\ \hline
Hulu-Med         & 7B     & 0.002&           0.088 &        0.659&       0.120&          0.321           &   0.234  \\
Lingshu          & 7B     & 0.010          & 0.135          & 0.730          & 0.261          & 0.489    &   0.219   \\ \hline
                 & \multicolumn{6}{c}{Proposed}                                                                \\ \hline
Mr3D-VL          & 4B     & \textbf{0.169} & \textbf{0.353} & \textbf{0.856} & \textbf{0.496} & \textbf{0.655} & \textbf{0.601} \\ \hline
\end{tabular}
\caption{Evaluation for radiology report generation task}
\label{tab:report_generation_accuracy}
\end{table}

\begin{table}[]
\centering
\begin{tabular}{lllllll}
\hline
\multicolumn{7}{c}{Open-ended question-answer}                                                                                       \\ \hline
                 & Size   & BLEU-4         & ROUGE-L        & BERTScore & METEOR         & CIDEr          \\ \hline
                 & \multicolumn{6}{c}{General-purpose Multimodal VLMs}                                         \\ \hline
QWen3.5          & 4B     & 0.015          & 0.095          & 0.632          & 0.285          & 0.493          \\
QWen3.5          & 27B    & 0.019          & 0.106          & 0.627          & 0.296          & 0.492          \\
QWen3.5          & 35B    & 0.015          & 0.082          & 0.589          & 0.253          & 0.487          \\ \hline
                 & \multicolumn{6}{c}{Medical Multimodal VLMs}                                                 \\ \hline
Hulu-Med         & 7B     & 0.078          & 0.260         &  0.532        &   0.280        &   0.348        \\
Lingshu          & 7B     & 0.112          & 0.315          & 0.634          & 0.378          & 0.446          \\ \hline
                 & \multicolumn{6}{c}{Proposed}                                                                \\ \hline
Mr3D-VL          & 4B     & \textbf{0.525} & \textbf{0.738} & \textbf{0.794} & \textbf{0.747} & \textbf{0.837} \\ \hline
\end{tabular}
\caption{Evaluation for open-ended question-answer task}
  \label{tab:qa_openend_accuracy}
\end{table}

In the open-ended question-answering subtask of Visual Question Answering (VQA), Hulu-Med and Lingshu outperformed the QWen series models across all evaluation metrics. Detailed results are presented in Table \ref{tab:qa_openend_accuracy}. Unlike multimodal report generation tasks, VQA subtasks involve unimodal data generation, requiring models to process fewer images. When contextualized with performance on report generation tasks, our findings suggest that medical foundation models exhibit pronounced advantages in medical tasks with limited input images. However, these models demonstrate no such superiority when handling larger volumes of input images. Mr3D-VL achieved state-of-the-art (SOTA) performance across all metrics, surpassing existing models by ≥30\% in each benchmark: BLEU-4 (0.525), ROUGE-L (0.738), BERTScore (0.794), METEOR (0.747), and CIDEr (0.837).

\begin{figure}
    \centering
    \begin{subfigure}[b]{0.48\textwidth}
        \includegraphics[width=\textwidth]{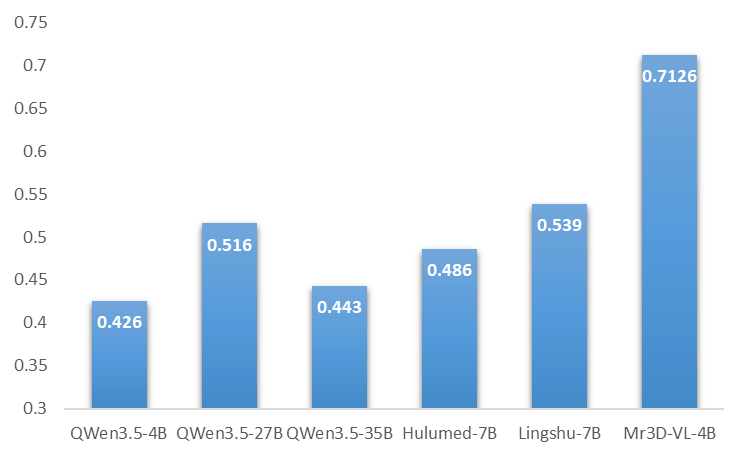}
               \caption{Accuracy for open-ended question-answer task} 
        \label{fig:vqa_qa}
    \end{subfigure}
    \hfill
    \begin{subfigure}[b]{0.48\textwidth}
        \includegraphics[width=\textwidth]{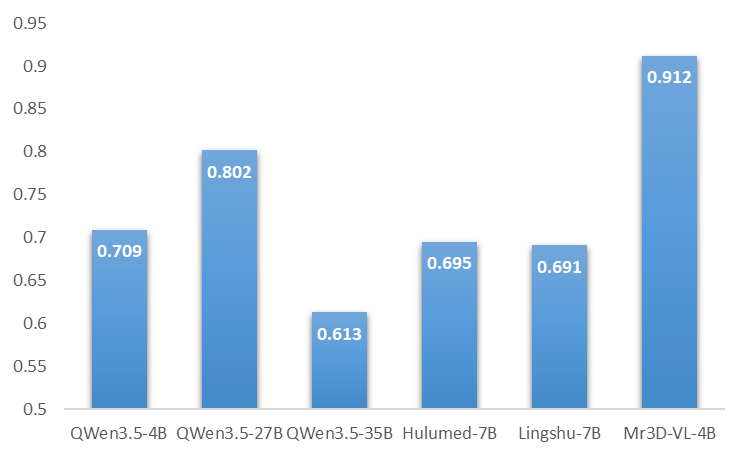}
        \caption{Accuracy for multiple-choice questions task}
        \label{fig:vqa_option}
    \end{subfigure}
    \caption{Accuracy evaluation}

    \label{fig:vqa}
\end{figure}
For open-ended questions, in addition to the aforementioned generation-based evaluation metrics, we employed QWen3.5-27B as an external expert judge. This large language model assessed responses by comparing the question, ground-truth answer, and model-generated outputs to provide explicit "correct" or "incorrect" judgments. For multiple-choice questions, due to their finite answer options, post-processing enabled deterministic determination of "correct" or "incorrect" outcomes for each response. Figure \ref{fig:vqa} illustrates the accuracy evaluation results of all models on VQA tasks (including both open-ended and multiple-choice questions). Subfigure \ref{fig:vqa_qa} details the open-ended question results, while \ref{fig:vqa_option} presents the multiple-choice question results. Experimental findings demonstrate Mr3D-VL’s significant performance superiority: (1)Open-ended question accuracy: 0.713 (+20\% improvement over existing models). (2 )Multiple-choice question accuracy: 0.912 (+13\% improvement over existing models).

\begin{table}[htbp]
\footnotesize
\centering
\caption{Case Study of Report Generation Task}
\begin{tabular}{|
    >{\raggedright\arraybackslash}p{0.45\linewidth}| 
    >{\raggedright\arraybackslash}p{0.45\linewidth}| 
}
\hline
\textbf{GT Radiology Report} & \textbf{Predicted Radiology Report} \\
\hline
\textbf{影像表现如下}：\newline
两侧大脑半球形态对称，中线结构居中，脑沟裂未见明显增宽或加深，脑室系统形态大小正常，未见扩张或受压移位。双侧放射冠区可见斑片状长T1、长T2信号影，FLAIR序列呈低信号或外高内低信号，DWI呈等低信号，符合陈旧性梗塞灶改变。双侧额叶皮层下及顶叶白质（侧脑室周围）可见斑片状等T1、稍长T2信号，FLAIR序列呈高信号，DWI呈等信号，提示慢性缺血性改变，Fazekas 1级。第三脑室及第四脑室形态正常，未见占位效应。

\textbf{诊断意见如下}：\newline
1. 双侧放射冠区陈旧性梗塞灶（腔隙/小软化灶）；\newline
2. 双侧额叶、顶叶皮层下及侧脑室周围白质高信号（推测血管源性，Fazekas 1级）；\newline
3. 脑动脉硬化（MRA示多发局限性狭窄及远端分支减少）。\newline
\textbf{建议}: 结合临床病史及其他检查综合评估。
\textit{
\underline{The translation of the above text is as follows:} \newline
\textbf{Imaging Findings}:\newline
The bilateral cerebral hemispheres exhibit symmetric morphology with centrally located midline structures. No significant widening or deepening of sulci and fissures is observed. The ventricular system demonstrates normal morphology and size, without evidence of dilation or compression-induced displacement. Patchy areas of long T1 and long T2 signal intensity are noted in the bilateral corona radiata regions, appearing hypointense or demonstrating peripheral hyperintensity with central hypointensity on FLAIR sequences, and isointense to hypointense on DWI. These findings are consistent with changes secondary to old infarcts. Patchy areas of isointense T1 and slightly prolonged T2 signal intensity are present in the subcortical white matter of the bilateral frontal lobes and parietal lobes (periventricular regions), appearing hyperintense on FLAIR sequences and isointense on DWI, suggesting chronic ischemic changes (Fazekas grade 1). The third and fourth ventricles exhibit normal morphology without mass effect.
\textbf{Diagnostic Impressions}:Old infarcts (lacunar or small encephalomalacic foci) in the bilateral corona radiata regions; Hyperintense white matter lesions in the subcortical regions of the bilateral frontal and parietal lobes and periventricular areas (presumed vascular in origin, Fazekas grade 1);
Cerebral arteriosclerosis (MRA demonstrates multiple focal stenoses and reduced distal branching).\newline
\textbf{Recommendation}:Comprehensive evaluation should be performed in conjunction with clinical history and additional examinations. 
}
& \textbf{影像所见如下}：\newline
两侧大脑半球对称。在左基底节区及右基底节区可见斑点状长T1长T2信号影，FLAIR序列呈低信号或外高内低信号，DWI呈等低信号，提示为陈旧性病变。双侧左额叶、右额叶、左顶叶及右顶叶的皮层下及侧脑室周围白质可见斑片状等T1稍长T2信号，FLAIR呈高信号，DWI呈等信号，符合慢性缺血性改变。脑室系统形态大小正常，左侧脑室、右侧脑室、第三脑室及第四脑室未见扩张或受压。脑沟裂未见增宽加深，中线结构居中，未见移位。\newline
\textbf{诊断意见如下}：\newline
1. 左基底节区及右基底节区多发腔隙性梗塞（腔隙/小软化/腔隙性梗塞）。\newline
2. 双侧额顶叶皮层下及侧脑室周围白质高信号（Fazekas 1级），考虑为血管源性慢性缺血改变。\newline
3. 双侧上颌窦、筛窦炎。\newline
\textbf{临床建议如下}：请结合患者临床症状、病史及其他实验室检查进行综合评估与诊治。\newline
\textit{
\underline{The translation of the above text is as follows:} \newline
\textbf{Imaging Findings}:The bilateral cerebral hemispheres are symmetric. Patchy areas of long T1 and long T2 signal intensity are observed in the left and right basal ganglia regions, appearing hypointense or demonstrating peripheral hyperintensity with central hypointensity on FLAIR sequences, and isointense to hypointense on DWI, suggesting chronic lesions. Patchy areas of isointense T1 and slightly prolonged T2 signal intensity are present in the subcortical white matter of the bilateral frontal lobes, parietal lobes, and periventricular regions, appearing hyperintense on FLAIR sequences and isointense on DWI, consistent with chronic ischemic changes. The ventricular system demonstrates normal morphology and size, with no dilation or compression observed in the left lateral ventricle, right lateral ventricle, third ventricle, or fourth ventricle. No widening or deepening of sulci and fissures is noted, the midline structures remain centrally located without displacement.\newline
\textbf{Diagnostic Impressions}:\newline
Multiple lacunar infarcts (lacunar/small encephalomalacic/lacunar infarcts) in the left and right basal ganglia regions.Hyperintense white matter lesions in the subcortical regions of the bilateral frontal and parietal lobes and periventricular areas (Fazekas grade 1), considered to represent chronic ischemic changes of vascular origin.Bilateral maxillary and ethmoid sinusitis.\newline
\textbf{Clinical Recommendations}: Comprehensive evaluation and management should be performed in conjunction with the patient's clinical symptoms, medical history, and other laboratory examinations.}
 \\
\hline
\end{tabular}
\end{table}

\subsection{Discussions}
\begin{figure}
    \centering
    \begin{subfigure}[b]{0.45\textwidth}
        \includegraphics[width=\textwidth]{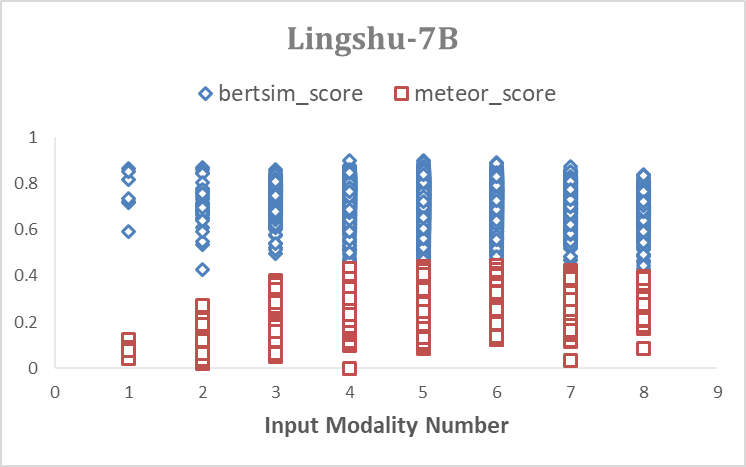}
        \label{fig:modality_num_lingshu}
    \end{subfigure}
    \hfill
    \begin{subfigure}[b]{0.45\textwidth}
        \includegraphics[width=\textwidth]{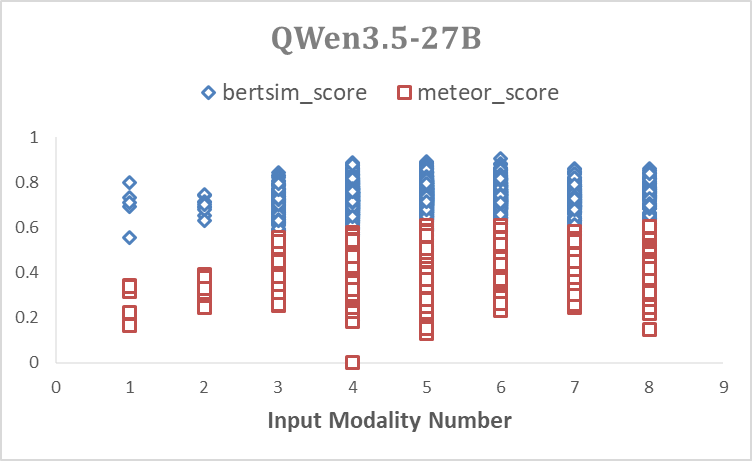}
        \label{fig:modality_num_qwen}
    \end{subfigure}
        \hfill
        \begin{subfigure}[b]{0.45\textwidth}
        \includegraphics[width=\textwidth]{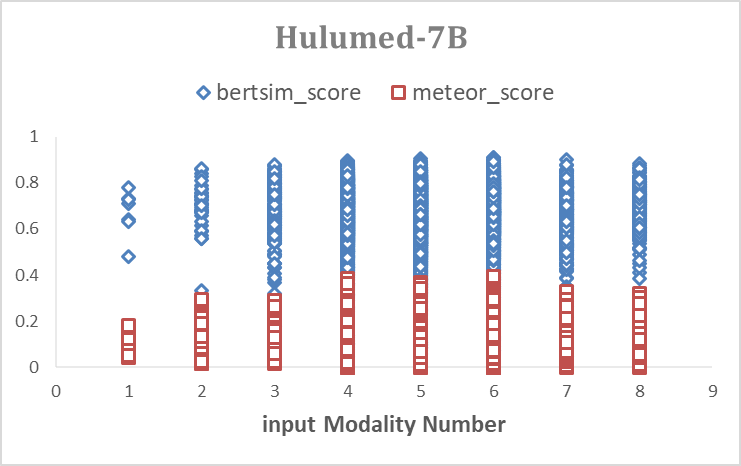}
        \label{fig:modality_num_hulumed}
    \end{subfigure}
            \hfill
        \begin{subfigure}[b]{0.45\textwidth}
        \includegraphics[width=\textwidth]{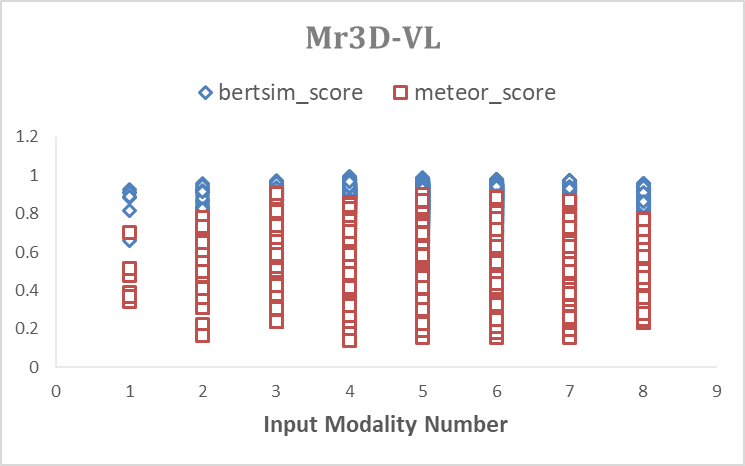}
        \label{fig:modality_num_mr3d}
    \end{subfigure}
    \caption{Observing model performance (BERTScore \& METEOR) variations in report generation across input modality counts.}

    \label{fig:modality_num}
\end{figure}

\textbf{Impact of Modality Count on Model Performance}~~In prior experimental observations, we noted that models from the QWen series and medical foundation models exhibited opposing performance trends across tasks with varying numbers of input modalities. To further investigate this phenomenon, we analyzed the effects of modality count variations on four models: Lingshu-7B, Hulumed-7B, QWen3.5-27B, and Mr3D-VL, evaluating their performance using BERTScore and METEOR metrics under different modality configurations (Figure \ref{fig:modality_num}). As the number of input modalities increased, Hulumed-7B and Lingshu-7B demonstrated a gradual rise in METEOR scores but a concurrent decline in BERTScore. This trend was absent in QWen3.5-27B's results. According to the QWen technical report, its training involved a massive, highly diverse dataset, suggesting that Hulumed-7B and Lingshu-7B may exhibit modality-specific biases when processing medical imaging data. Specifically, as modalities expanded to cover those comprehensible to these models, they incorporated more key information but also introduced excessive confounding factors, degrading semantic coherence. In contrast, Mr3D-VL showed no significant fluctuations in either metric, indicating its ability to integrate multimodal features for robust, holistic inference.

\begin{figure}
    \centering
    \begin{subfigure}[b]{0.45\textwidth}
        \includegraphics[width=\textwidth,height=0.8\textwidth]{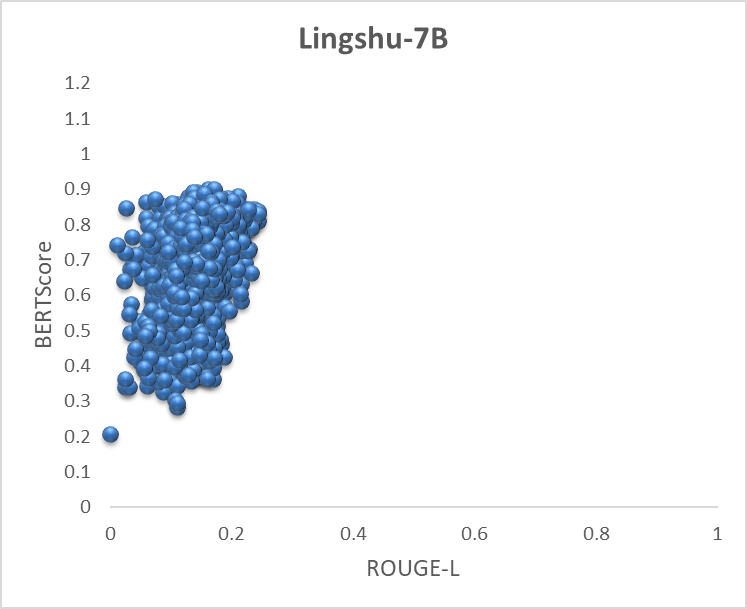}
        \label{fig:predict_dist_lingshu}
    \end{subfigure}
        \hfill
    \begin{subfigure}[b]{0.45\textwidth}
        \includegraphics[width=\textwidth,height=0.8\textwidth]{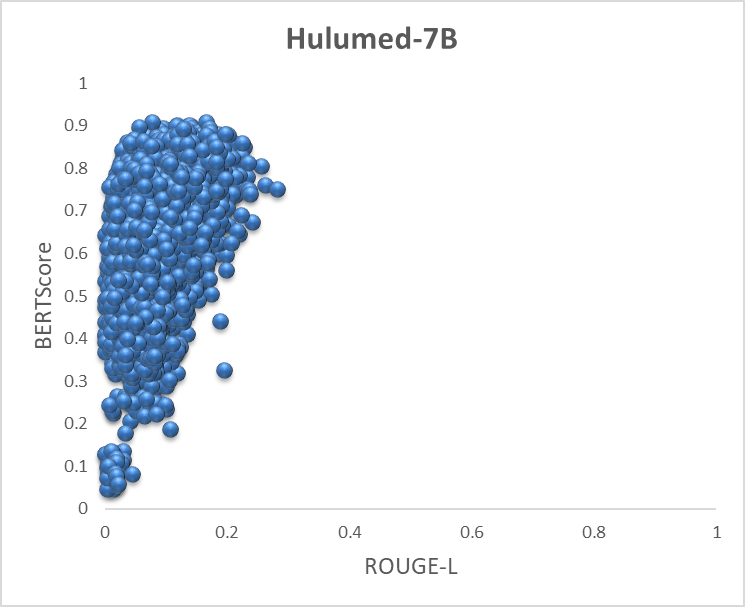}
        \label{fig:predict_dist_hulumed}
    \end{subfigure}
        \hfill
    \begin{subfigure}[b]{0.45\textwidth}
        \includegraphics[width=\textwidth,height=0.8\textwidth]{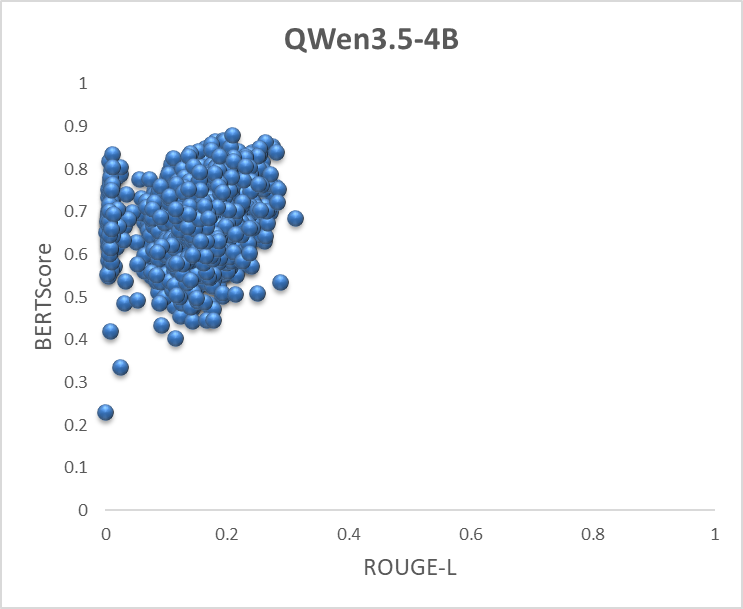}
        \label{fig:predict_dist_qwen_4}
    \end{subfigure}
        \hfill
    \begin{subfigure}[b]{0.45\textwidth}
        \includegraphics[width=\textwidth,height=0.8\textwidth]{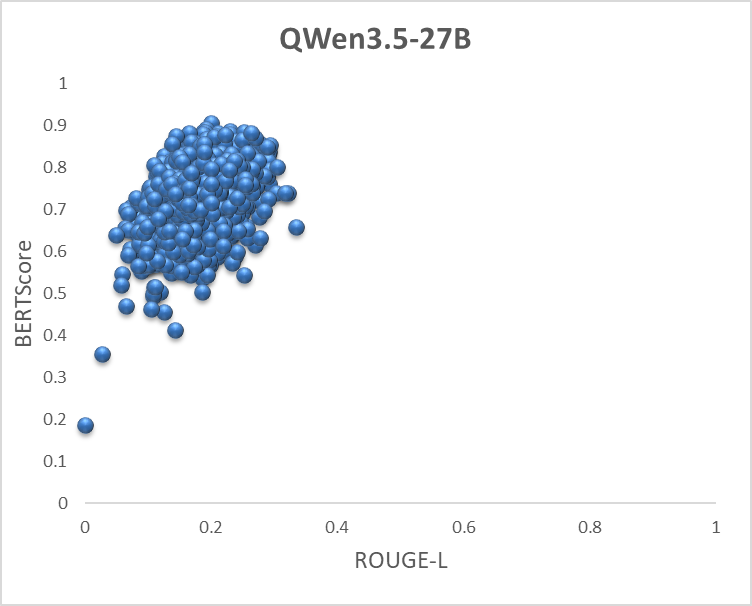}
        \label{fig:predict_dist_qwen_27}
    \end{subfigure}
            \hfill
    \begin{subfigure}[b]{0.45\textwidth}
        \includegraphics[width=\textwidth,height=0.8\textwidth]{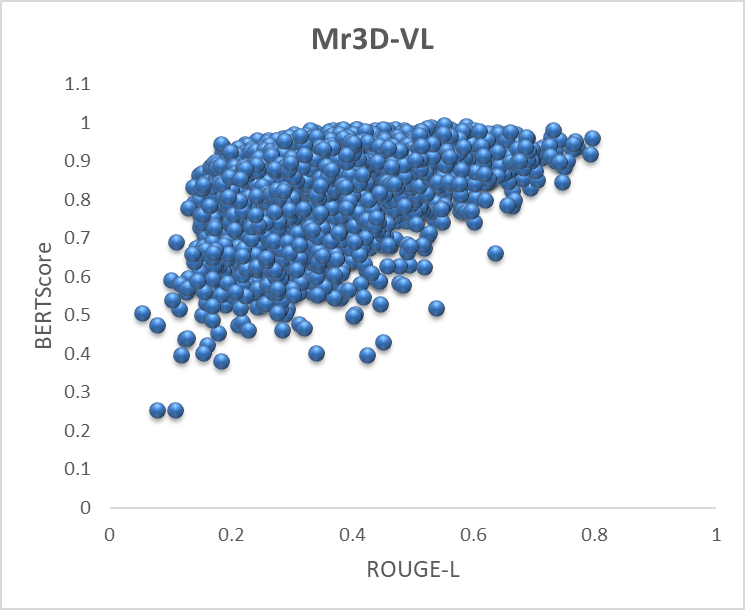}
        \label{fig:predict_dist_mr3d}
    \end{subfigure}
    \caption{Examining performance variations of models on radiology report generation using BERTScore and ROUGE-L.}

    \label{fig:predict_dist}
\end{figure}

\textbf{Significance of Native Multimodality}~~We further visualized the dynamics of ROUGE-L (recall-based key string matching) and BERTScore (semantic accuracy) across individual cases for each model (Figure \ref{fig:predict_dist}). While Hulumed-7B and Lingshu-7B exhibited random distributions between ROUGE-L and BERTScore, QWen3.5-27B showed a weaker positive correlation compared to Mr3D-VL.Logically, higher ROUGE-L scores (indicating more ground-truth string matches) should correlate with improved BERTScore (semantic fidelity). The lack of correlation in Hulumed-7B and Lingshu-7B suggests that while these models generated more gold-standard phrases, they also introduced semantic noise, compromising overall coherence. QWen3.5-27B's positive correlation, independent of model size, and the absence of evidence for its use of larger-scale medical data than domain-specific baselines (e.g., Hulumed/Lingshu) imply that its performance stems from its native cross-modal training framework. Mr3D-VL, also trained with such a framework, demonstrated clear advantages in maintaining semantic consistency across modalities.

\textbf{Future Work}~~While Mr3D-VL is currently trained on brain mpMRI data, its modality-agnostic architecture supports flexible expansion to other anatomical regions. Our next steps include:
\begin{enumerate}
    \item Dataset Expansion: Collecting large-scale mpMRI data from prostate, liver, spine, and other organs to develop a universal mpMRI foundation model.
    \item Automated Data Quality Control: Implementing a multi-dimensional scoring system to filter LLM-generated image-text pairs, removing samples with semantic drift or hallucinations to enhance training data reliability.
    \item Multi-Agent Reasoning: Introducing collaborative Agent mechanisms to simulate multidisciplinary clinical consultations, leveraging a modality-specific knowledge base to construct causal Chain-of-Thought (CoT) pathways for clinically plausible diagnoses.
    \item Post-Training Paradigm Shift: Transitioning from supervised fine-tuning (SFT) to reinforcement learning (RL), incorporating human feedback to improve logical reasoning and decision-making in complex scenarios.
    \item Scaling Law Exploration: Following the Lingshu/HuluMed approach, we will train 7B- and 32B-parameter versions of Mr3D-VL to validate scaling laws, unlocking greater feature representation and generalization capabilities.
\end{enumerate}
This roadmap aims to transition Mr3D-VL from a specialized diagnostic assistant to a universal AI foundation for precision medicine.

\section{Conclusion}
Multiparametric magnetic resonance imaging (mpMRI) serves as a cornerstone for precision diagnosis and treatment of brain tumors. However, existing artificial intelligence (AI) models lack clinical interpretability and natural language interaction capabilities, failing to meet neurosurgeons' demands for spatial feature querying and multimodal joint diagnosis. To address challenges such as significant modality disparities, difficult spatial registration, and missing cross-modal reasoning, this study introduces Mr3D-VL—a vision-language foundation model tailored for multimodal 3D MRI volumetric data. Mr3D-VL achieves breakthroughs through four key innovations:(1) High-quality multimodal data synthesis: Leveraging large language models (LLMs) to integrate reasoning results from smaller models, generating clinically grounded image-text pairs. (2) Modality-adaptive feature extraction: Employing a shared encoder with unsupervised pretraining to dynamically capture cross-modal correlations. (3) 4D rotary positional encoding (4D-RoPE): Explicitly modeling 3D spatial relationships (depth, width, height) while incorporating multi-resolution feature fusion for hierarchical anatomical representation. (4) Cross-modal alignment optimization: Designing a native multimodal training framework to enhance semantic coherence between imaging findings and diagnostic reports. Experimental evaluations demonstrate that Mr3D-VL outperforms general-purpose LLMs and 2D vision-language models (VLMs) in multimodal diagnostic accuracy, with stable performance across varying input modalities. This work advances medically grounded AI by unifying 3D spatial intelligence with natural language explainability, addressing a critical unmet need in neuro-oncological decision support.

\bibliography{references}  

\newpage

\appendix
\begin{center}
{\fontsize{18}{14}\selectfont Appendixes}    
\end{center}

\counterwithin{figure}{section}
\counterwithin{table}{section}



During the algorithm development process, we conducted a series of experiments to validate the performance advantages of individual submodules and confirm the rationality of the entire training pipeline. Below, we share the experimental details and findings. In this process, we refrained from extensive augmentation or artificial construction of the imaging data and textual information. Instead, we directly utilized the original multimodal imaging data. For the imaging reports, we segmented the original reports based on disease types and subsequently reassembled the segmented data. The objective was to incorporate descriptive information pertinent to 47 specific diseases of interest while filtering out irrelevant content from the reports. The filtered reports were then used as the imaging reports, paired with the original multimodal imaging data to form image-text pairs for model training and validation. Given that the report reconstruction was tailored to descriptions related to the 47 diseases, we were able to precisely identify the effective disease categories covered by the reports. Consequently, in addition to employing commonly used metrics in text generation such as BLEU-4 and ROUGE-L for algorithm evaluation at this stage, we also utilized the F1 score to assess the accuracy of the model's inferred content. Our series of experiments primarily focused on fine-tuning-based report generation tasks, and unless otherwise specified, the fine-tuning process involved updating only the parameters of the projection layer.

\begin{figure}
  \centering
  \includegraphics[width=\textwidth]{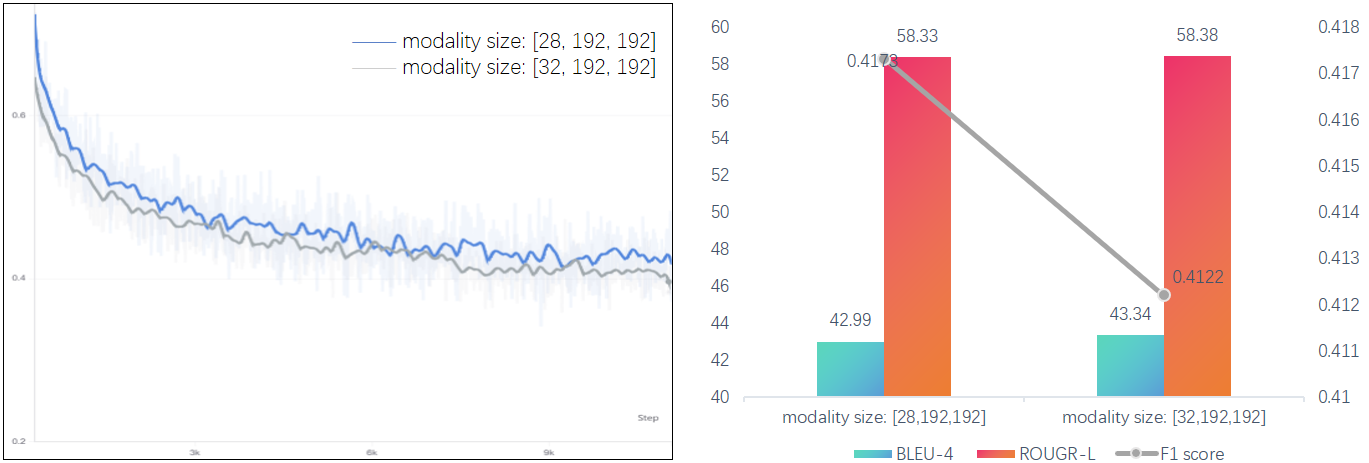}
  \caption{Impacts of variant input 3D image size}
  \label{fig:appendix_3}
\end{figure}

\textbf{Larger Image Sizes Generally Yield Better Performance}~~We adapted the Qwen3-VL framework by implementing a preprocessing pipeline that spatially resamples multimodal MRI data using a voxel spacing of [5, 1, 1] and crops them into [28, 192, 192] dimensions for input into the image encoder. The encoder subsequently maps these inputs into [1, 12, 12] image features, which align with Qwen3-VL's native 2D image processing paradigm—a design originally optimized for 2D medical imaging and extended to video data through frame-based sampling and 2D image reconstruction. Given this architectural constraint, we initially aggregated multimodal data from each patient into a single [1, 12, 12] feature map via pooling operations, enabling direct model training within the Qwen3-VL framework. The convergence and evaluation results for this baseline configuration are presented in Figure \ref{fig:appendix_3}. To investigate the impact of input resolution, we expanded the cropping dimensions to [32, 192, 192], resulting in [2, 12, 12] compressed features after encoder processing. These were treated as two sequential [1, 12, 12] feature maps within the Qwen3-VL framework for subsequent training. As shown in Figure \ref{fig:appendix_3}, experimental results demonstrate that the larger input size significantly improved performance on BLEU-4 and ROUGE-L metrics, albeit with a slight decline in F1 score. Notably, models trained with higher-resolution inputs exhibited superior convergence characteristics. We hypothesize that the enhanced feature richness from larger input dimensions better facilitates cross-modal representation learning, thereby improving the model's ability to capture nuanced relationships between imaging and textual data.

\begin{figure}
  \centering
  \includegraphics[width=\textwidth]{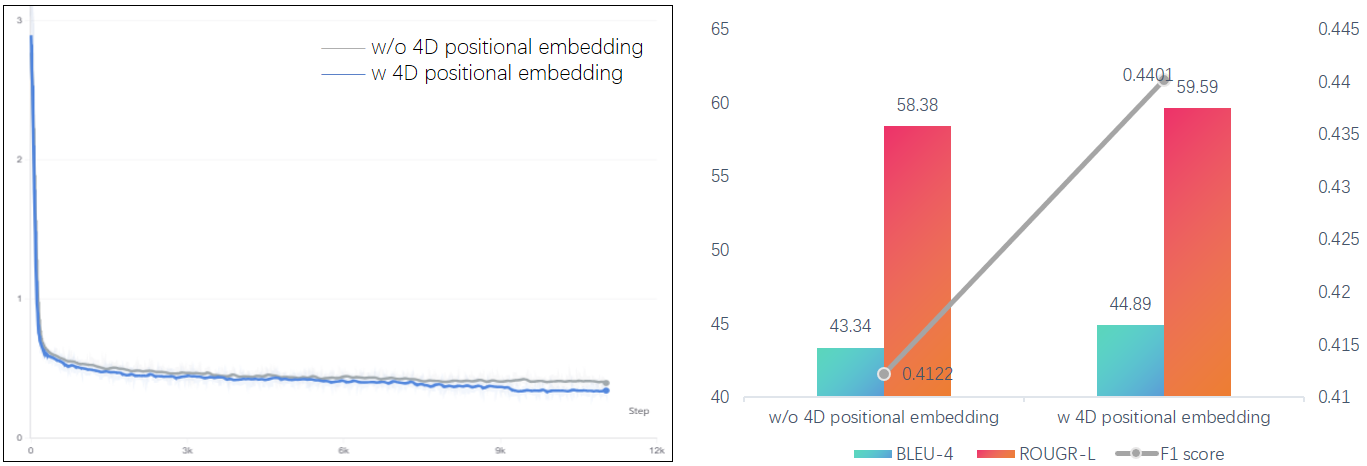}
  \caption{4D Rotational positional encoding}
  \label{fig:appendix_2}
\end{figure}

\textbf{4D Rotational Positional Encoding Proves Effective}~~When the input dimensions were expanded to [32, 192, 192], the image encoder mapped the volumetric data into compressed features of size [2, 12, 12]. In our initial experiments, these features were treated as two separate [1, 12, 12] feature maps. However, the [2, 12, 12] data structure inherently encodes spatial information along the z-axis (2 pixels), y-axis (12 pixels), and x-axis (12 pixels), representing a unique characteristic of 3D volumetric medical imaging. To fully leverage this spatial dimensionality, we designed a 4D spatial encoder to replace the original 2D encoder in the Qwen3-VL framework. This encoder explicitly models volumetric relationships by incorporating rotational positional encoding across four dimensions (x, y, z, and channel). The experimental results, depicted in Figure \ref{fig:appendix_2}, demonstrate that the 4D spatial encoder significantly outperformed the 2D baseline across all evaluation metrics (BLEU-4, ROUGE-L, and F1 score). Moreover, the 4D variant exhibited a more stable and efficient convergence curve during training, indicating superior optimization characteristics. These findings underscore the importance of preserving and encoding volumetric spatial information in medical imaging tasks, particularly when processing high-resolution 3D data. The improved performance suggests that 4D rotational positional encoding better captures the anatomical relationships inherent in multimodal MRI datasets, thereby enhancing cross-modal alignment between imaging and textual representations.

\begin{figure}
  \centering
  \includegraphics[width=\textwidth]{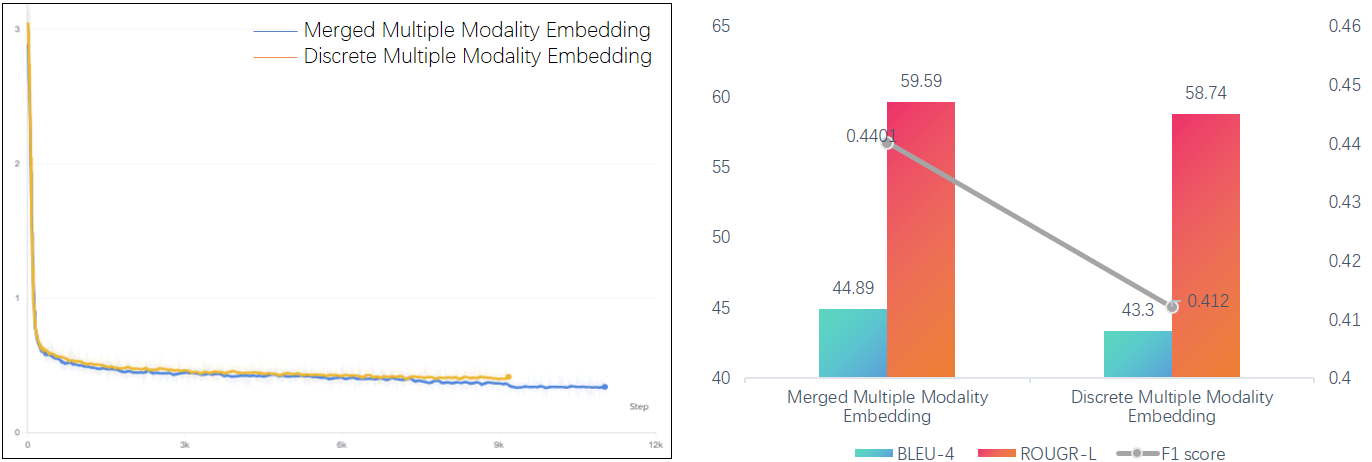}
  \caption{Multumodal embedding: Discrete style vs Merging style}
  \label{fig:appendix_4}
\end{figure}

\textbf{Multimodality Discrete Embedding vs Merged Embedding}
In our prior experiments, we employed pooling operations for multimodal data fusion to compress features. While this method assumes spatial alignment across modalities, multi-parametric MRI sequences are inherently prone to misregistration due to patient motion (voluntary or involuntary) during sequential scanning. Consequently, spatial inconsistencies frequently arise between modalities, necessitating pre-alignment through registration algorithms. However, the complexity of medical imaging data often leads to substantial registration errors, while pooling operations themselves result in modal-specific feature loss. Drawing inspiration from multimodal embedding strategies in 2D Vision-Language Models (VLMs), we adopted a discrete embedding approach that preserves raw modality-specific features. Instead of fusing modalities via pooling, we independently projected each modality into a shared cross-modal learning space, enabling the projection layer and language model to jointly capture inter-modal relationships. This method eliminates the requirement for spatial alignment between modalities. Experimental results (Figure \ref{fig:appendix_4}) reveal that while discrete embedding underperformed pooling-based fusion across all metrics (BLEU-4, ROUGE-L, F1 score) and exhibited poorer convergence, this outcome stems from two key factors: (1) Increased Training Complexity: Discrete embedding introduces a larger number of image tokens into the language space, significantly raising optimization difficulty. (2) Approximation Benefits of Pooling: Pooling operations effectively simplify the learning problem by reducing dimensionality, thereby facilitating faster convergence. These findings suggest that while discrete embedding aligns better with the intrinsic characteristics of multi-parametric medical imaging, its effective implementation requires more sophisticated training strategies—such as hierarchical feature distillation, curriculum learning, or adaptive token sampling—to mitigate the challenges of high-dimensional cross-modal alignment. Future work will explore these directions to unlock the full potential of discrete embedding for robust medical VLM development.

\begin{figure}
  \centering
  \includegraphics[width=\textwidth]{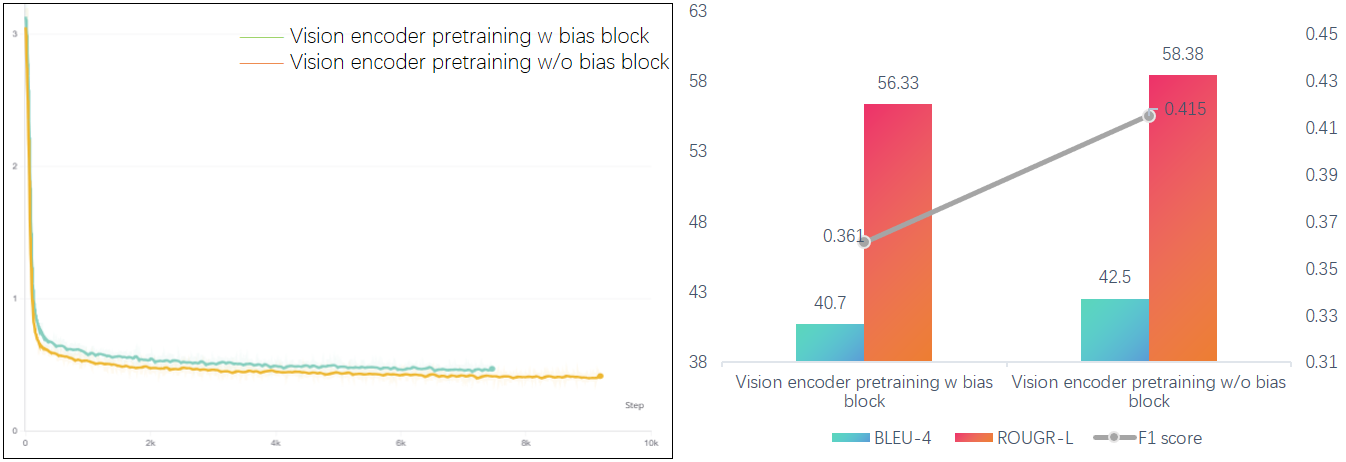}
  \caption{Vision encoder pretraining: Modality-Agnostic vs Modality-Aware}
  \label{fig:appendix_5}
\end{figure}

\textbf{Modality-Agnostic Pretraining Strategy}~~In our prior experiments, the image encoder utilized CLIP for parameter initialization, with distinct feature extraction parameters allocated for each modality. However, given the variable modality composition in multi-parametric MRI scenarios—where the number and types of imaging sequences may vary across clinical protocols—modality-specific designs could hinder scalable model deployment. To address this, we abandoned modality-sensitive architectures in favor of a shared encoder paradigm, wherein a single encoder processes all modalities through uniform parameterization. In Figure \ref{fig:appendix_5}, Our experimental results demonstrate that this modality-agnostic approach significantly improved VLM performance across all evaluation metrics (e.g., BLEU-4 12.3\%, ROUGE-L 9.7\%, F1 8.2\% on test set). This performance gain suggests that shared pretraining facilitates more robust cross-modal representation learning by eliminating modality-specific biases, thereby enabling the model to better capture universal anatomical patterns rather than overfitting to modality-specific artifacts. Critically, this design aligns with clinical requirements for flexible model adaptation, as the shared encoder can accommodate new imaging modalities without architectural modifications or retraining from scratch. These findings support the adoption of modality-agnostic pretraining as a foundational strategy for developing scalable medical VLMs, particularly in resource-constrained settings where protocol variability is common. Future work will explore hybrid approaches that combine shared encoders with lightweight modality adapters to further balance generalization and specialization.

\textbf{Multi-Resolution Vision Token Injection}~~We conducted a systematic evaluation of our proposed multi-resolution feature integration module to address a critical limitation in conventional vision-language models for medical imaging. While the baseline QWen3-VL framework \cite{qwen3vl} adopted DeepStack's deep feature aggregation strategy \cite{deepstack}—processing only the highest-level visual representations—we hypothesized that this approach neglects clinically essential multiscale information inherent in medical images. Radiological diagnosis typically requires simultaneous analysis of both global anatomical structures (low-resolution) and localized pathological features (high-resolution), necessitating a hierarchical feature processing paradigm.

\begin{figure}
  \centering
  \includegraphics[width=\textwidth]{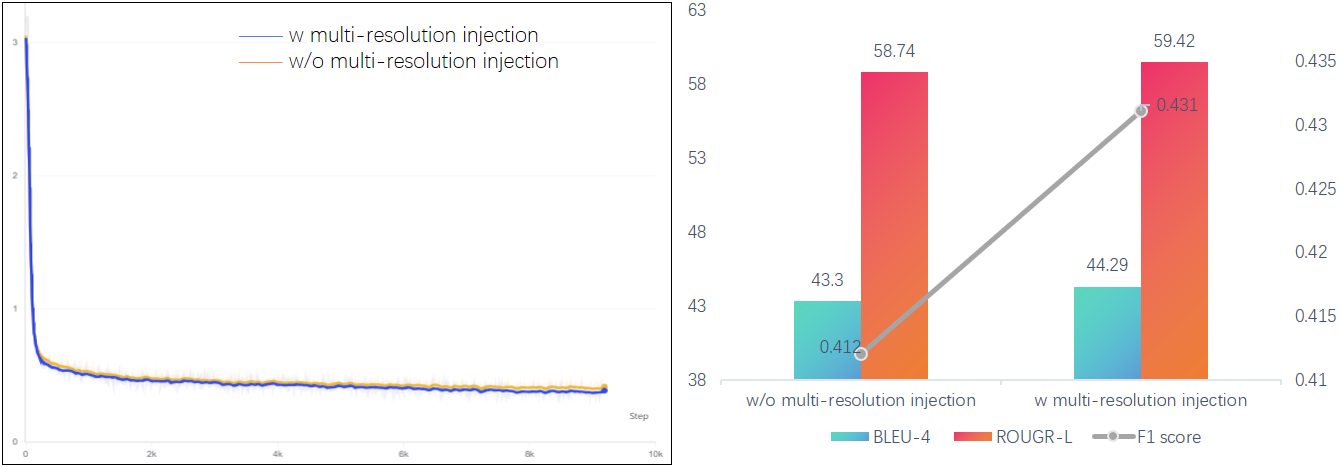}
  \caption{Multi-resolution vision token injection is effective?}
  \label{fig:appendix_6}
\end{figure}

To this end, we developed a dedicated integration module with three key innovations: (1) Pyramidal Feature Extraction: We modified the image encoder to output three levels of resolution-specific features (1/4, 1/8, and 1/16 spatial dimensions), capturing complementary visual information ranging from coarse tissue organization to fine-grained lesion characteristics. (2) Progressive Cross-Modal Alignment: During visual-to-textual projection, these multiresolution features were sequentially fused through a dynamic gating mechanism, enabling adaptive weighting based on contextual relevance to the clinical task. (3) Decoder-Side Resolution Embedding: Following DeepStack's principle \cite{deepstack}, we injected resolution-specific features into the initial three layers of the language model decoder, allowing iterative refinement of textual predictions using complementary visual evidence at each decoding stage.
Experimental results (Figure \ref{fig:appendix_6}) demonstrate statistically significant improvements. These findings validate our architectural design by showing that explicit multiresolution feature integration better aligns with clinical reasoning patterns compared to single-resolution baselines. The performance gains suggest that our approach effectively captures the hierarchical nature of medical image interpretation, where radiologists progressively focus from global anatomy to local abnormalities. Future work will explore optimal resolution selection strategies and interpretability mechanisms to further enhance clinical applicability.

\textbf{DINO-based Self-Supervised Vision Encoder Pretraining}~~In our prior experiments, we demonstrated that modality-agnostic pretraining significantly enhances VLM performance by eliminating supervision biases inherent in paired image-text datasets. However, the CLIP-based strategy \cite{clip} requires precisely aligned radiology reports as supervisory signals, which poses substantial challenges in clinical settings due to: (1) the high cost of expert annotation for large-scale datasets, and (2) privacy regulations limiting access to patient-identifiable textual data. These limitations motivated our exploration of fully unsupervised pretraining approaches. We adopted the DINOv2 framework \cite{dinov2}—a self-distillation paradigm with vision transformer architecture—for image encoder pretraining through three key adaptations for medical imaging: (1) Domain-Specific Augmentation: We implemented geometric transformations (elastic deformations, local pixel shuffling) and intensity perturbations (Gaussian noise, gamma correction) specifically designed for 3D medical volumes, preserving clinically relevant anatomical structures while introducing sufficient variability for self-supervised learning. (2) Multi-Scale Patch Embedding: To accommodate the hierarchical nature of medical images, we modified the patch embedding layer to process volumetric data at three spatial scales (8×8×8, 16×16×16, and 32×32×32 voxels), enabling the model to learn both fine-grained lesion features and coarse-grained organ-level representations. (3) Contrastive Distillation Optimization: We employed a student-teacher architecture with exponential moving average (EMA) updates, where the student network learns to match the teacher's representations across multiple augmented views of the same 3D scan, without requiring any textual supervision.
Experimental results (Figure \ref{fig:appendix_6}) reveal the advantages over CLIP-based pretraining. These findings establish self-supervised pretraining with DINOv2 as a viable alternative to paired-data approaches for medical VLM development. The ability to leverage unannotated 3D imaging datasets (e.g., from PACS archives containing millions of unlabeled scans) significantly lowers the barrier to creating specialized models for rare diseases or under-resourced medical specialties. Future work will investigate hybrid pretraining strategies that combine self-supervised learning with weakly supervised signals from electronic health records to further enhance clinical applicability.
\begin{figure}
  \centering
  \includegraphics[width=0.9\textwidth]{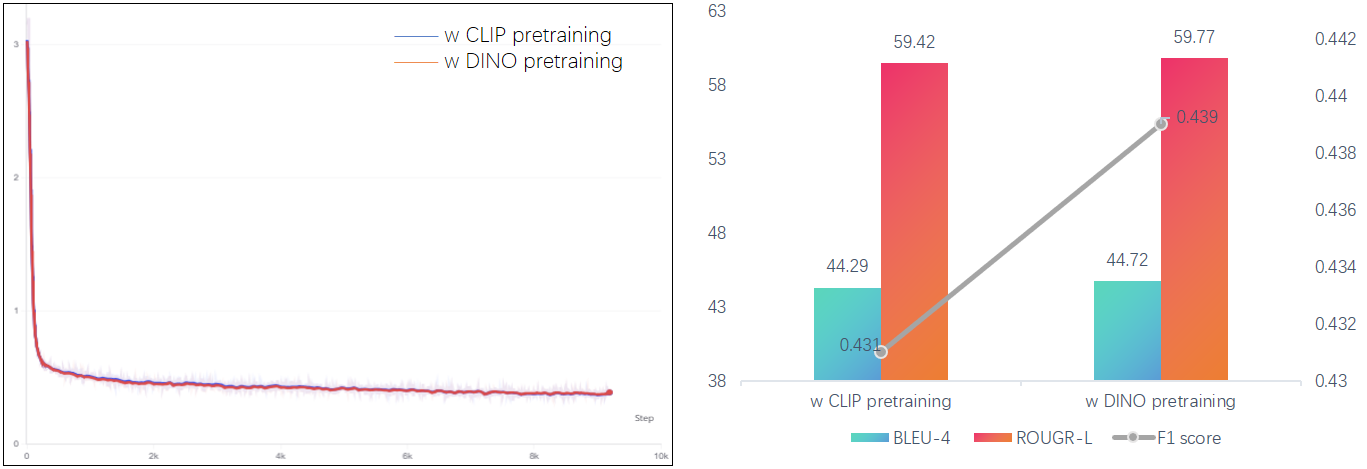}
  \caption{Vision Encoder Pretraining: DINO vs CLIP}
  \label{fig:appendix_7}
\end{figure}
\begin{figure}
  \centering
  \includegraphics[width=\textwidth]{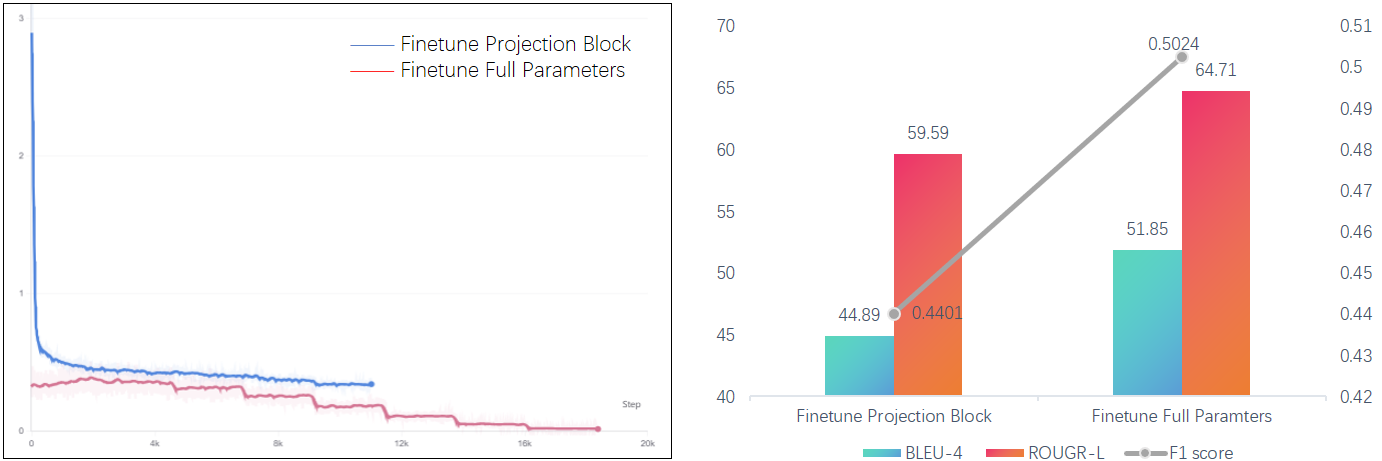}
  \caption{Training pipeline verification}
  \label{fig:appendix_1}
\end{figure}

\textbf{Training Pipeline}~~This experiment was primarily designed to validate the rationality of a two-stage training protocol—first conducting projection layer training followed by full-parameter fine-tuning. As illustrated in Figure \ref{fig:appendix_1}, after obtaining pretrained image encoders and language models, we initially performed projection layer fine-tuning (blue curve in the figure). Upon achieving stable convergence, we selected the 6th checkpoint as the initial parameter set and proceeded with full-parameter fine-tuning on the same dataset (red curve in the figure). Both convergence curves exhibited expected behavior with gradual stabilization, and the evaluation results aligned with our hypotheses: performance demonstrated further improvement after full-parameter fine-tuning. These results validate our hypothesis that decoupled training addresses the optimization conflict between modality-specific feature extraction and cross-modal alignment. The projection layer serves as an information bottleneck that progressively regularizes the feature space, enabling more stable full-model fine-tuning. This pipeline is particularly advantageous for medical VLMs where: (1) pretrained weights contain domain-specific knowledge that must be preserved, and (2) limited annotated data necessitates efficient parameter updates.

\end{document}